\documentclass[pdflatex,sn-mathphys-num]{sn-jnl}% Math and Physical Sciences Numbered Reference Style
\usepackage{graphicx}%
\usepackage{multirow}%
\usepackage{amsmath,amssymb,amsfonts}%
\usepackage{amsthm}%
\usepackage{mathrsfs}%
\usepackage[title]{appendix}%
\usepackage{xcolor}%
\usepackage{textcomp}%
\usepackage{manyfoot}%
\usepackage{booktabs}%
\usepackage{algorithm}%
\usepackage{algorithmicx}%
\usepackage{algpseudocode}%
\usepackage{listings}%
\usepackage{graphicx}
\usepackage{caption}
\theoremstyle{thmstyleone}%
\theoremstyle{thmstyletwo}%

\theoremstyle{thmstylethree}%

\begin{document}

\title[Article Title]{Dynamic Weight Adjustment for Knowledge Distillation: Leveraging Vision Transformers for High-Accuracy Lung Cancer Detection and Real-Time Deployment}


\author*[1,3]{\fnm{Saif Ur Rehman} \sur{Khan}}\email{saif\_ur\_rehman.khan@dfki.de}
%xuz72wot@rptu.de
%saif\_ur\_rehman.khan@dfki.de
\author*[1,2]{\fnm{Muhammad Nabeel} \sur{Asim}}\email{muhammad\_nabeel.asim@dfki.de}
\author[1,2]{\fnm{Sebastian} \sur{ Vollmer}}\email{sebastian.vollmer@dfki.de}
\author[1,2,3]{\fnm{Andreas} \sur{Dengel}}\email{andreas.dengel@dfki.de}

\affil[1]{\orgdiv{German Research Center for Artificial Intelligence}, \orgaddress{ \city{Kaiserslautern}, \postcode{67663}, \country{Germany}}}
\affil[2]{\orgdiv{Intelligentx GmbH (intelligentx.com)}, \orgaddress{ \city{Kaiserslautern}, \country{Germany}}}
\affil[3]{\orgdiv{Department of Computer Science}, \orgname{Rhineland-Palatinate Technical University of Kaiserslautern-Landau} \orgaddress{ \city{Kaiserslautern}, \postcode{67663}, \country{Germany}}}


\abstract{This paper presents the FuzzyDistillViT-MobileNet model, a novel approach for lung cancer (LC) classification, leveraging dynamic fuzzy logic-driven knowledge distillation (KD) to address uncertainty and complexity in disease diagnosis. Unlike traditional models that rely on static KD with fixed weights, our method dynamically adjusts the distillation weight using fuzzy logic, enabling the student model to focus on high-confidence regions while reducing attention to ambiguous areas. This dynamic adjustment improves the model ability to handle varying uncertainty levels across different regions of LC images. We employ the Vision Transformer (ViT-B32) as the instructor model, which effectively transfers knowledge to the student model, MobileNet, enhancing the student generalization capabilities. The training process is further optimized using a dynamic wait adjustment mechanism that adapts the training procedure for improved convergence and performance. To enhance image quality, we introduce pixel-level image fusion improvement techniques such as Gamma correction and Histogram Equalization. The processed images (Pix1 and Pix2) are fused using a wavelet-based fusion method to improve image resolution and feature preservation. This fusion method uses the wavedec2 function to standardize images to a 224x224 resolution, decompose them into multi-scale frequency components, and recursively average coefficients at each level for better feature representation. To address computational efficiency, Genetic Algorithm (GA) is used to select the most suitable pre-trained student model from a pool of 12 candidates, balancing model performance with computational cost. The model is evaluated on two datasets, including LC25000 histopathological images (99.16\% accuracy) and IQOTH/NCCD CT-scan images (99.54\% accuracy), demonstrating robustness across different imaging domains. Interpretability is ensured with GRAD-CAM, GRAD-CAM++, and LIME for visualizing the regions that the model focuses on during predictions. Finally, an Android application is developed for real-time deployment, confirming the model practical applicability for medical professionals.}




\maketitle
\begin{table}[h!]
\centering
\begin{tabular}{cc}
\hline
\textbf{Abbreviations} & \textbf{Full form} \\
\hline
LC & Lung cancer \\

KD & Knowledge distillation \\

ViT & Vision transformer \\

GA & Genetic algorithm \\

NSCLC & Non-small cell lung \\

SCLC & Small cell lung cancer \\

CNN & Convolutional neural networks \\

DL & Deep learning \\

KL & Kullback-Leibler \\
\hline
\end{tabular}
\end{table}

\section{Introduction}\label{sec1}

LC is the leading cause of cancer-related deaths globally. Early detection and accurate classification of LC are crucial for improving patient outcomes and treatment. It results from uncontrolled growth of cells in the lungs by forming malignant tumors that can spread. LC is categorized into two major histological subtypes like non-small cell lung cancer (NSCLC) and small cell lung cancer (SCLC) \cite{Yang2024}.  NSCLC is the most prevalent form by accounting for 85\% of all LC and consisting of three distinct subtypes such as adenocarcinoma, squamous cell carcinoma, and large cell carcinoma \cite{Romeo2023}. In contrast, SCLC is less common that accounts for approximately 13-15\% of all LC cases but characterized by its  aggressive nature, rapid proliferation, and early spread, with exceptionally high growth rates, a strong tendency for early metastasis, and most patients are diagnosed with metastatic disease \cite{Wang2022}. Symptoms of LC include chronic coughing, chest pain, bone or spinal discomfort, unintended weight loss, and fatigue may manifest several months prior to formal diagnosis \cite{Prado2023}. Medical Imaging methods can be categorized into non-invasive and invasive techniques \cite{Gupta2024}. Non-invasive techniques, such as chest X-rays, computed tomography (CT), positron emission tomography (PET), and magnetic resonance imaging (MRI) have been extensively utilized for the early detection of LC \cite{Wang2022b}. Chest X-rays detect lung nodules, CT scans provide detailed cross-sectional images revealing the size, shape, and location of affected areas, MRIs help to assess the extent of cancer and its spread to other tissues, and PET scans identify cancer cells based on their increased metabolic activity \cite{Alif2025}. Although these conventional imaging techniques are widely used, the early detection of LC continues to be challenging due to the subtle nature of clinical symptoms and inherent limitations in manual analysis. These challenges highlight the importance of developing automated diagnostic systems to enhance diagnostic precision of LC.

The emergence of artificial intelligence (AI) \cite{Svoboda2020}, particularly through Deep Learning (DL) techniques, has proven to be a transformative advancement in improving diagnostic precision of LC. DL \cite{Rajasekar2023,Wani2024}, which leverages Convolutional Neural Networks (CNN), is increasingly utilized for the analysis of large datasets and medical imaging for enabling the detection of complex patterns and subtle features within LC images. CNN \cite{Humayun2022,Atiya2024} models can autonomously learn from raw image data to identify more intricate patterns within medical images that are often difficult to detect using traditional methods, ultimately contribute to improved patient outcomes. DL models often require substantial computational resources and large datasets for training, which can limit their practical deployment in real-time or resource-constrained clinical environments. This challenge underscores the importance of KD \cite{ElAssiouti2024,Elbatel2023}, which is a technique that enables the transfer of knowledge from a large and complex model (teacher) to a smaller more efficient model (student) without sacrificing too much performance. By leveraging KD, the smaller student model can maintain high accuracy while being computationally efficient by making it more suitable for deployment in clinical practice. Traditional KD generates soft labels using fixed weight for assuming uniform importance across all image regions. However, this approach overlooks varying uncertainty levels in medical images. Fewer prior studies \cite{Pavel2024,Li2024,Zheng2023,Tian2024} employed traditional KD methods to enhance model efficiency for LC diagnosis but did not incorporate dynamic fuzzy weighting to adaptively focus on high-confidence regions. A more effective strategy incorporates dynamic weighting with fuzzy scale weights for enabling the model to focus on high-confidence regions while down-weighting ambiguous areas. Fuzzy logic assigns overlapping confidence values by allowing for more flexible and advanced decision-making, ultimately improving diagnostic accuracy. 

\subsection*{Problem statement} 
In traditional KD, soft labels are calculated using a fixed weight W, balancing the cross-entropy loss and Kullback-Leibler (KL) divergence. While effective in many cases, this approach fails when applied to data with varying uncertainty, particularly in complex disease image analysis. In such images, certain regions may have well-defined features, while others are noisy or ambiguous, requiring more nuanced interpretation. The fixed weight W in conventional KD treats all regions equally, neglecting areas with higher uncertainty that need more focused learning. This leads to poor performance in regions with greater complexity, especially in disease images where precise learning is critical. The conventional KD loss function is given by:
\begin{equation}
L_{\text{KD}} = \rho \times \text{CrossEntropy}_L (Y - \hat{Y}) + (1 - \rho) \times A^2 \times \text{KL}(\lambda_{\text{Soft}}(A), \lambda_{\text{hard}})
\end{equation}
This fixed weighting approach does not effectively address the complexity inherent in disease image analysis, where some regions may require more focused learning due to higher uncertainty. Consequently, there is a need for a more dynamic approach that adjusts the loss function based on the varying levels of uncertainty across different regions of the image, enabling the model to focus more on areas that require higher attention and improving overall performance. \\
\subsection*{Novelty of this study} 
Our approach introduces dynamic weighting through fuzzy scale weights to address varying uncertainty levels in disease images, allowing the model to adaptively focus on high-confidence regions while down-weighting ambiguous areas. Instead of treating the image uniformly, fuzzy logic assigns overlapping confidence values to different regions, enabling a more flexible and nuanced decision-making process. Rather than using a fixed scaling factor, we propose a dynamic weight $\mathbf{\omega}(x)$ that modulates the contribution of the KD loss based on local uncertainty in the image. This allows the model to adjust the instructor influence according to the confidence or uncertainty at different regions.
\begin{equation}
\omega(x) = \mu_{\text{low}}(x) \times w_{\text{low}} + \mu_{\text{medium}}(x) \times w_{\text{medium}} + \mu_{\text{high}}(x) \times w_{\text{high}} \tag{2}
\end{equation}
Where: 
\begin{itemize}
    \item $\mu_{\text{low}}(x)$, $\mu_{\text{medium}}(x)$, and $\mu_{\text{high}}(x)$ are the fuzzy membership values for low, medium, and high confidence.
    \item $w_{\text{low}}$, $w_{\text{medium}}$, and $w_{\text{high}}$ are the corresponding weights for each confidence level.
\end{itemize}
The main contribution of this work as follows:
\begin{itemize}
    \item \textbf{Learn Feature with Vision Transformer:} This work employs the ViT-B32 as the instructor model for feature learning, effectively transferring knowledge to the student model. The transfer process is enhanced by a dynamic wait adjustment mechanism that adapts the training procedure, improving model convergence and performance. ViT's ability to capture long-range dependencies and provide global contextual information offers a significant advantage over traditional CNN-based models, enhancing the student model's generalization capabilities.
    \item \textbf{Dynamic Weight Adjustment using Fuzzy Logic:} We introduce a dynamic weight adjustment mechanism using fuzzy logic, offering an improvement over conventional KD techniques that rely on static weights. This approach enables the student model to adaptively focus on high-confidence regions while reducing attention on ambiguous areas, improving performance and the model's ability to manage varying levels of uncertainty.
    \item \textbf{High-Accuracy Evaluation on Diverse Datasets:} The model is rigorously evaluated on two diverse datasets: LC25000 histopathological images, achieving an accuracy of 99.16\%, and IQOTH/NCCD CT-scan images, with an accuracy of 99.54\%. These results showcase the model's robustness and its ability to generalize across different medical imaging domains, demonstrating its potential for real-world applications.
    \item \textbf{Interpretability and Real-Time Deployment:} Efficient interpretability methods, including GRAD-CAM, GRAD-CAM++, and LIME, are employed to visualize and analyze the regions of focus during predictions, ensuring transparency in the model's decision-making process. Additionally, an Android application is developed for real-time deployment, confirming the model's practical applicability and effectiveness in a mobile environment.
\end{itemize}
\section{Related work}
Several methods have been developed to improve diagnostic accuracy in LC detection with a growing emphasis on improving computational efficiency for real-time clinical applications. Among them, Pavel et al.[15] proposed a three-stage KD framework with a teaching assistant for detecting NSCLC using CT scan images. Their method utilizes a ViT as the teacher, ResNet152V2 as the teaching assistant, and a custom CNN as the student model. The teaching assistant model achieved a test accuracy of 90.99\%, while the student model reached 94.53\%. However, their method is computationally intensive due to reliance on a three-stage training process by incorporating deep pretrained architectures. Zheng et al. \cite{Zheng2023} introduced KD-ConvNeXt, a teacher-student network architecture leveraging KD for classifying lung tumors from histopathological images. Their method allows the student network (ConvNeXt) to extract knowledge from the intermediate feature layers of the teacher network (Swin Transformer) for enhanced feature extraction and achieved a classification accuracy of 85.64\%. However, their approach is limited by its relatively moderate accuracy compared to more recent methods and the dependence on a large teacher model during training. Tian et al.\cite{Tian2024} developed a DL model for diagnosing LC by combining a Feature Pyramid Network (FPN), Squeeze-and-Excitation (SE) modules, and ResNet18 architecture. The performance was further enhanced through KD by transferring knowledge from larger teacher models to more compact student models. Their model achieved an average accuracy of 98.84\% using a histopathology LC dataset. Shariff et al. \cite{Shariff2025} introduced a CNN model with Differential Augmentation (DA) to address memory overfitting, a key limitation that affects the generalization of models to unseen data in LC detection. Their approach incorporates targeted augmentation strategies to enhance data diversity and improve model robustness. The authors achieved an accuracy of 98.78\% by testing on IQ-OTH/NCCD LC dataset. However, computational efficiency and accuracy could be further improved by applying advanced fuzzy-based dynamic KD techniques. Pathan et al. \cite{Pathan2024} developed an optimized CNN architecture for LC screening by leveraging the Sine Cosine Algorithm (SCA) for hyperparameter optimization. The CNN model consisting of five convolutional layers utilizes SCA to minimize the error rate by optimizing tuning parameters as an objective function. The authors achieved an impressive 99\% average classification accuracy for distinguishing between normal, benign, and malignant lung scans. Priya et al. \cite{Priya2025} proposed a DL-based model for LC classification by employing the SE-ResNeXt-50-CNN architecture. The model was tested on the IQ-OTHNCCD dataset and achieved an impressive accuracy of 99.15\%. Their approach integrates dynamic Quadri-histogram equalization (QDHE)-based preprocessing, data augmentation, and hyperparameter optimization to enhance model performance. However, despite its high accuracy, the computational cost of their model poses a challenge due to the reliance on deep pretrained architectures. Akter et al. \cite{Akter2021} proposed an algorithm that integrates fuzzy-based image segmentation to enhance the segmentation of lung nodules in CT images. The authors employ a neuro-fuzzy classifier to distinguish between malignant and benign nodules by exhibiting a classification accuracy of 90\%. Their approach demonstrates considerable promise in improving the early detection of LC. Yan et al. \cite{Yan2023} developed a CNN model for the automated detection of LC in CT images. The CT images were preprocessed prior to input into the CNN, and the model was further optimized using a modified Snake Optimization algorithm (SOA). Their proposed model achieved a classification accuracy of 96.58\% when evaluated on the IQ-OTH/NCCD-LC dataset. The accuracy performance of their model can be further enhanced by incorporating more advanced techniques, such as dynamic KD approach. Zhang et al.\cite{Zhang2023} developed a DL model that uses ResNet combined with a Convolutional Block Attention Module (CBAM) to classify LC as benign or malignant using CT images. Their model achieved test accuracy of 89.80\%, but its performance can be improved with the application of more advanced approaches. All the methods discussed above employed different approaches to diagnose LC. While some utilized traditional KD techniques but dynamic KD approach remains a challenging task that offers potential for further improvement in diagnostic accuracy. Table 1 summarizes the various KD-based methods for lung cancer detection, highlighting their objectives, performance, and limitations as discussed in the related works \cite{Pavel2024,Zheng2023,Tian2024,Shariff2025,Pathan2024,Priya2025}.
\begin{table}[h!]
\centering
\caption{Summary of KD Approaches for LC Detection}
\begin{tabular}{p{2.5cm}p{3cm}p{3cm}p{3cm}p{2.8cm}}
\hline
\textbf{Reference} & \textbf{Method} & \textbf{Objective} & \textbf{Performance} & \textbf{Limitation} \\
\hline
Pavel et al. \cite{Pavel2024} & Three-stage KD framework with ViT (teacher), ResNet152V2 (assistant), custom CNN (student) & NSCLC detection using CT scan images & Student model achieved 94.53\% accuracy & Computationally intensive due to three-stage training and reliance on deep pretrained architectures \\

Zheng et al. \cite{Zheng2023} & KD\_ConvNeXt (teacher-student network) & Lung tumor classification from histopathological images & Achieved 85.64\% classification accuracy & Moderate accuracy and dependence on a large teacher model during training \\

Tian et al. \cite{Tian2024} & DL model with FPN, SE modules, ResNet18, enhanced by KD & LC diagnosis with histopathology dataset & Achieved 98.84\% average accuracy & Relatively complex, requires larger teacher models for knowledge transfer \\

Shariff et al. \cite{Shariff2025} & CNN with Differential Augmentation (DA) & Improve model robustness and generalization in LC detection & Achieved 98.78\% accuracy using IQ-OTH/NCCD LC dataset & Computational efficiency and accuracy could be improved using advanced fuzzy-based dynamic KD techniques \\

Pathan et al. \cite{Pathan2024} & CNN optimized with Sine Cosine Algorithm (SCA) & LC screening with optimized CNN architecture & Achieved 99\% classification accuracy & High accuracy, but limited by the complexity of optimization methods and hyperparameters \\

Priya et al. \cite{Priya2025} & SE-ResNeXt-50-CNN with QDHE-based preprocessing and data augmentation & LC classification using IQ-OTH/NCCD dataset & Achieved 99.15\% accuracy & High computational cost due to reliance on deep pretrained architectures \\
\hline
\end{tabular}
\end{table}
\section{Proposed method}
In this section, we outline the key components of the proposed method for KD using fuzzy scaled weights. Section 3.1 details the datasets used for model training, followed by Section 3.2, where we describe the process of selecting a student model from a pool of CNN. In Section 3.3, we introduce the instructor model, which guides the distillation process, and in Section 3.4, we discuss the chosen student model for knowledge transfer. Section 3.5 addresses the limitations of traditional KD losses, highlighting areas for improvement. Section 3.6 presents the optimization approach using fuzzy scaled weights to enhance the KD loss function. In Section 3.7, we provide an overview of the proposed FuzzyDistillViT-MobileNet model, emphasizing the application of scaled weights in the distillation process. Finally, Section 3.8 outlines the experimental settings and hyperparameters used in the study, while Section 3.9 defines the evaluation metrics for assessing the performance of the FuzzyDistillViT-MobileNet model.
\subsection{Dataset detail}
The IQOTH/NCCD Kaggle dataset \cite{Alyasriy2021}, which was gathered over three months in late 2019, is made publicly available for use in this work. Expert radiologists and oncologists have annotated the dataset, which consists of CT scan images of LC patients and healthy people at different phases of the disease. It has 1,097 pictures of 110 patients divided into three groups: benign, normal, and malignant. For analysis, the original DICOM pictures were transformed to JPEG format. Fig 1 shows the dataset overall samples of the three classes following, which aids in visualizing the differences between benign, normal, and malignant states. Histopathological image of LC, which are openly accessible on Kaggle, make up the second dataset employed in this work \cite{Borkowski2019}. Three different classes of LC are included in the collection, with 5,000 samples in each class. Fig. describes the total dataset sample for each class such as lung-aca, lung-n, and lung-scc.
\begin{figure}[h!]
    \centering
    \includegraphics[width = 9cm]{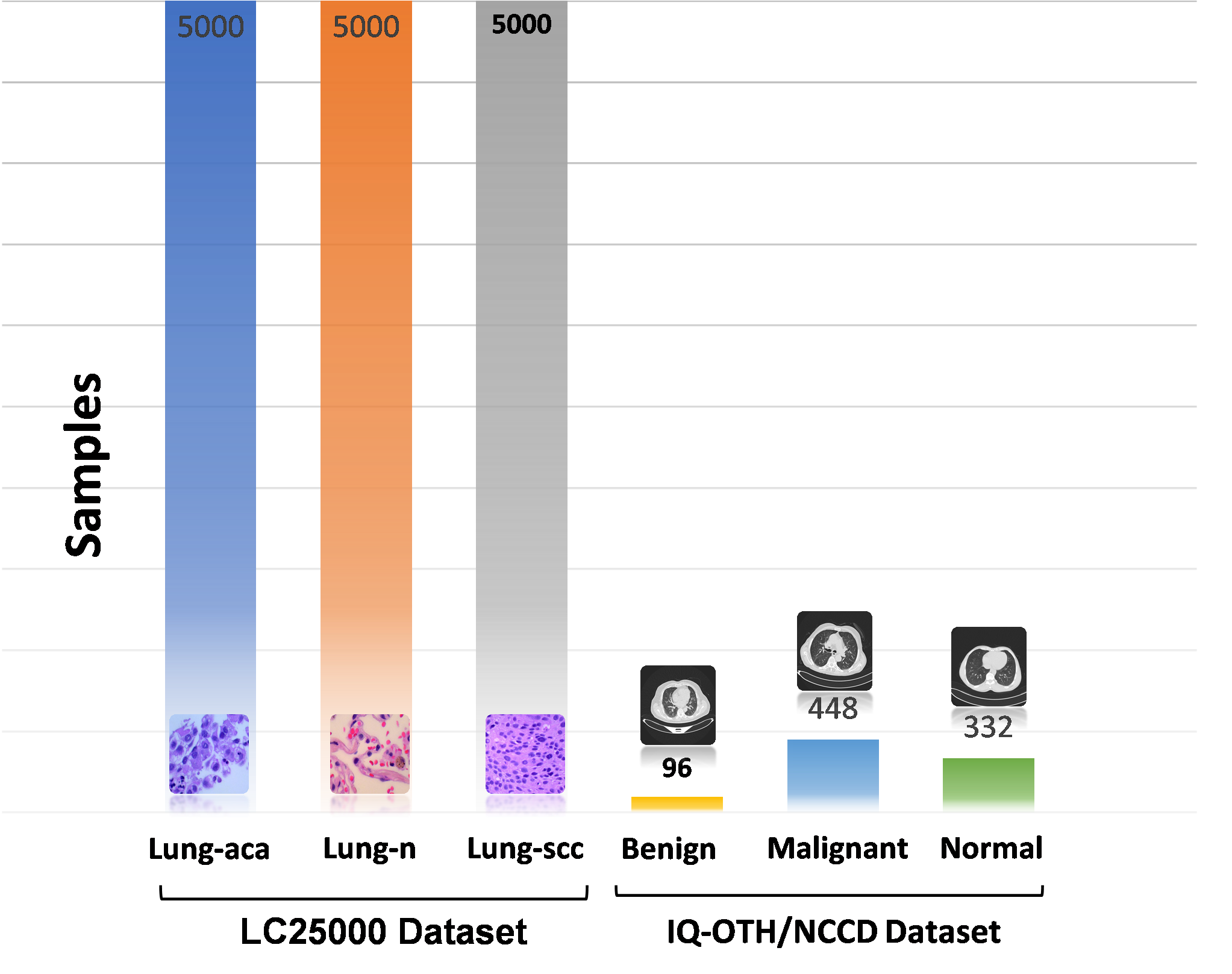}
    \caption{Dataset sample views of LC25000 dataset and IQ-OTH.NCCD dataset}
    \label{fig:se.png}
\end{figure}
\subsection{Selection of student model from Pool of CNN models}
In medical image classification, particularly for the diagnosis of LC, selecting an appropriate pre-trained model is essential for achieving accurate results. However, pre-trained models are trained on general datasets like ImageNet, which do not capture the complex features of specifically LC disease by leading to suboptimal performance. To overcome this, the GA was used to evaluate and select the most suitable pre-trained model. GA operates by simulating natural evolutionary processes, such as selection, crossover, and mutation, to improve a population of candidate models iteratively. Through this process, GA identifies the model that best captures the relevant features for LC detection, thus ensuring high classification accuracy while maintaining computational efficiency. In this study, GA successfully selected MobileNet as the optimal student model for this task. The methodology and mathematical formulation of GA are explained below by illustrating how it was applied to identify the most effective pre-trained model for the detection of LC.

\subsubsection*{Step-1: Population Initialization}
In Genetic Algorithm (GA), the initial population consists of individuals or chromosomes, each representing a potential solution to the problem. These individuals are randomly generated at the beginning of the algorithm, with each individual encoded as a vector of decision variables (genes). The initial population is denoted as \( \text{Pop}(0) \), where:
\begin{equation}
\text{Pop}(0) = \left\{ Y_1^{(0)}, Y_2^{(0)}, \dots, Y_M^{(0)} \right\} \tag{3}
\end{equation}
Here, \( Y_i^{(0)} \) represents the \( i \)-th individual in generation 0, and \( M \) is the total number of individuals (population size). Each chromosome \( Y_i^{(0)} \) is represented as a vector of genes (gen).
\begin{equation}
Y_i^{(0)} = \left[ \text{gen}_{i1}^{(0)}, \text{gen}_{i2}^{(0)}, \dots, \text{gen}_{in}^{(0)} \right]
\tag{4}
\end{equation}
Where \( n \) is the number of decision variables.

\subsubsection*{Step-2: Fitness Calculation}
The fitness of each individual in the population is evaluated using a fitness function Fitness, which measures how well a given solution approximates the optimal solution. The fitness score of individual \( Y_i^{(t)} \) at generation \( t \) is calculated as follows: 
\begin{equation}
\text{Fitness}_i^{(t)} = \text{Fitness}(Y_i^{(t)}) \tag{5}
\end{equation}
Where \( Fitness_i^{(t)} \) is the fitness score of individual \( Y_i^{(t)} \). The goal is to either maximize or minimize the fitness function by depending on the problem:
\begin{equation}
\text{maxFitness}(Y), \, Y \in \Omega \quad \text{or} \quad \text{minFitness}(Y), \, Y \in \Omega
\tag{6}
\end{equation}
Where \( \Omega \) represents the feasible solution space.

\subsubsection*{Step-3: Selection Criteria}
Selection is performed based on fitness values to ensure that better-performing individuals are more likely to be chosen for reproduction. The probability  \( P(Y_i) \) of selecting individual \( Y_i \) is based on their fitness, and it is given as:
\begin{equation}
Y_i = \frac{\text{Fitness}_i}{\sum_{j=1}^{M} \text{Fitness}_j} \tag{7}
\end{equation}
Where \( \text{Fitness}_i \) is the fitness of individual \( Y_i \) and 
\[
\sum_{j=1}^{M} \text{Fitness}_j
\]
is the total fitness of the entire population. This process ensures that individuals with higher fitness values have a higher probability of being selected.

\subsubsection*{Step-4: Crossover (Recombination) Criteria}
Crossover involves combining two parent individuals to produce offspring by exchanging genetic material. This process simulates the exchange of genes between individuals during reproduction. Two selected individuals, \( Y_1 \) and \( Y_2 \), undergo crossover by resulting in new offspring \( Y^{'} \):
\begin{equation}
Y' = \text{Crossover}(Y_1, Y_2) \tag{8}
\end{equation}
Let \( Y_1^{(t)} \) and \( Y_2^{(t)} \) be two parent chromosomes at generation \( t \), each consisting of \( k \) genes:

\[
Y_1^{(t)} = [ \text{gen}_{11}^{(t)}, \text{gen}_{12}^{(t)}, \dots, \text{gen}_{1k}^{(t)} ] \tag{9}
\]

\[
Y_2^{(t)} = [ \text{gen}_{21}^{(t)}, \text{gen}_{22}^{(t)}, \dots, \text{gen}_{2k}^{(t)} ] \tag{10}
\]
Where \( \text{gen}_{ij}^{(t)} \) represents the \( j \)-th gene of the \( i \)-th parent at generation \( t \), and \( k \) is the total number of genes in a chromosome. A random crossover point \( q \) is selected within the chromosome length \( k \), ensuring that genes before \( q \) remain intact, while genes from \( q+1 \) onward are exchanged between the parent chromosomes. As a result, the offspring chromosomes at generation \( t+1 \) are formed by combining parts of the parents' genetic information. This step aims to combine the strengths of the selected parents and generate potentially better solutions. Multiple crossover points can be used, depending on the algorithm.
\subsubsection*{Step-5:  Mutation Criteria}
The GA iterates through the selection, crossover, and mutation processes until a predefined stopping condition is met. The stopping criteria could be based on several factors, such as the maximum number of generations, the convergence of the population, or the discovery of a solution that satisfies a predefined fitness threshold. When any of these conditions are met, the algorithm terminates and returns the best solution found which is the optimal student model (MobileNet) from a pool of pre-trained models.
Mathematically, the stopping condition can be represented as:
\begin{equation}
\text{if } t \geq T_{\text{max}} \text{ or } \Delta F < \lambda \text{ or } \text{fitness}(Y_i^{(t)}) \geq F_{\text{threshold}}, \text{ stop.} \tag{11}
\end{equation}

Where,

\begin{itemize}
    \item $t$ is the current generation,
    \item $T_{\text{max}}$ is the maximum number of generations,
    \item $\Delta F$ is the change in fitness between generations,
    \item $\lambda$ is a small value indicating minimal improvement,
    \item $F_{\text{threshold}}$ is the predefined fitness threshold.
\end{itemize}
Once the condition is satisfied, the algorithm halts and outputs the best solution found during the iterations. This ensures that the algorithm stops when the solution converges or reaches a satisfactory level of performance.
\subsection{Instructor model}
The ViT model is an architecture designed for image recognition tasks, and the ViT-B32 variant is a specific configuration of this model. The ViT architecture represents a shift from traditional CNN by leveraging self-attention mechanisms commonly used in natural language processing. The key idea behind ViT is to treat an image as a sequence of patches rather than a grid of pixels. In ViT-B32, the image is split into non-overlapping 32x32 pixel patches, each of which is flattened and embedded into a 768-dimensional vector. These patch embeddings, along with position embeddings that provide spatial information, are fed into a standard Transformer architecture. The Transformer consists of multi-head self-attention layers and feedforward neural networks, which allow the model to capture long-range dependencies and context within the image. The ViT-B32 model has 12 layers, 12 attention heads, and a hidden size of 768, offering a balance between computational efficiency and model capacity. The ViT-B32 as an instructor model in a KD approach provides several benefits. The ViT has shown excellent performance in image classification tasks, particularly on large-scale datasets such as ImageNet. Its ability to capture global context through self-attention makes it highly effective for learning fine-grained details in images. As an instructor model, ViT-B32 can guide a smaller, student model by transferring its knowledge, which helps improve the student model performance even with fewer parameters. The KD enables the student to learn from the soft targets produced by the instructor model, which contain more information than the hard labels. This allows the student model to generalize better, even on limited data, while maintaining a smaller footprint, making it more suitable for real-time applications or deployment in resource-constrained environments. Fig 2 presents the architecture overview of the Vit model.
\begin{figure}[h!]
    \centering
    \includegraphics[width = 9cm]{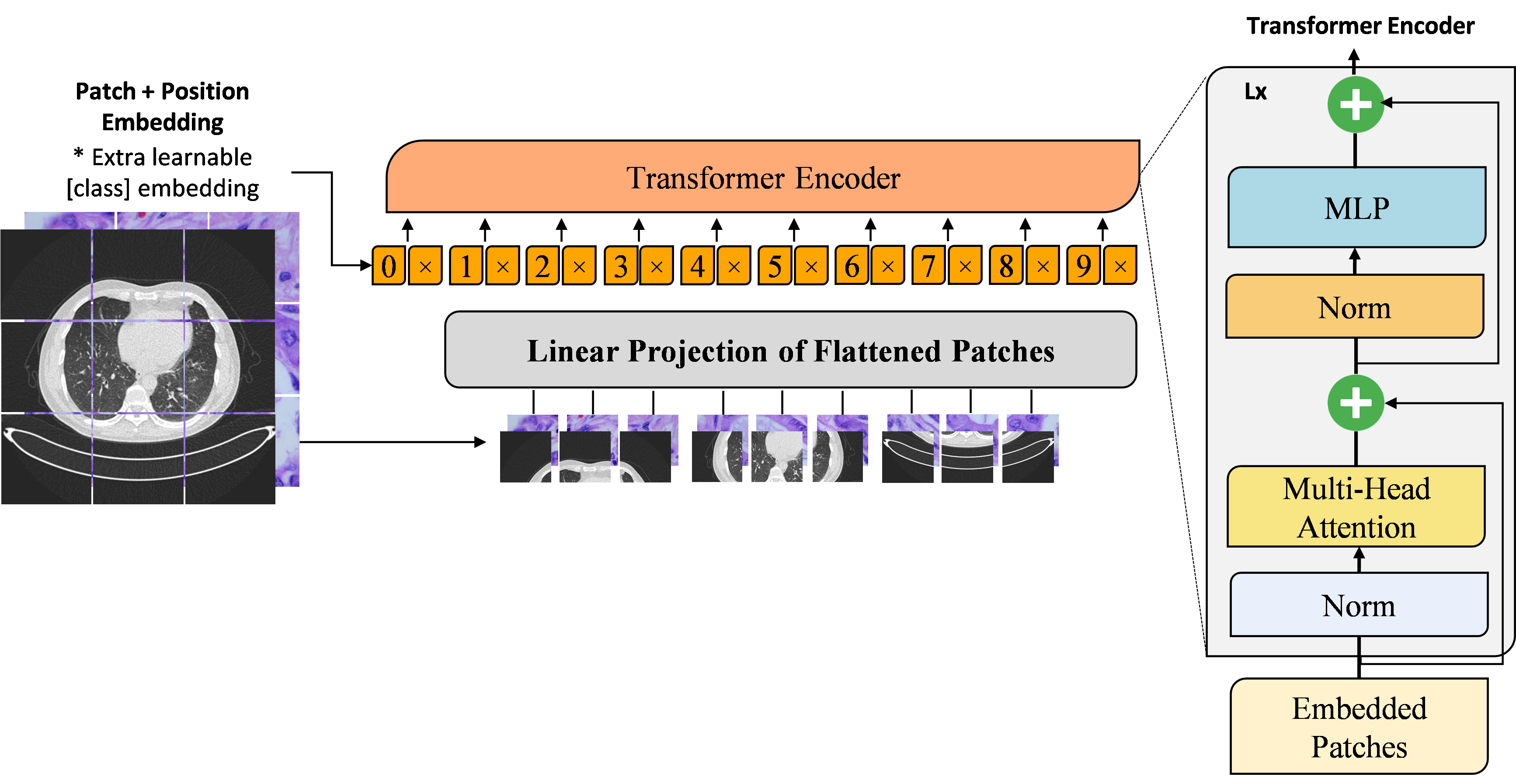}
    \caption{Architecture overview of transformer base instructor model}
    \label{fig:se.png}
\end{figure}
\subsection{Selected Student model by GA}
MobileNet is a lightweight DL model designed specifically for mobile and embedded vision applications, where computational resources and memory are limited. The core innovation behind MobileNet is the use of depthwise separable convolutions, which significantly reduce the number of parameters and computational cost compared to traditional convolutions. Depthwise separable convolutions break down a standard convolution into two smaller operations: a depthwise convolution, where each input channel is convolved separately, and a pointwise convolution, which is a 1x1 convolution that combines the outputs from the depthwise convolution. This decomposition results in a reduction in computation while maintaining competitive accuracy for image classification. MobileNet also employs the concept of an efficient depth multiplier and width multiplier, which allows the user to trade-off between accuracy and computational efficiency. The depth multiplier controls the number of channels in each layer, while the width multiplier adjusts the overall network width. These features enable the model to be scalable across various resource-constrained environments. MobileNet as a student model in a KD approach provides several benefits, particularly when optimizing for computational efficiency and model deployment in resource-constrained environments. KD involves transferring knowledge from a larger, more complex instructor model to a smaller student model, improving the student model's performance without requiring extensive retraining. Since MobileNet is inherently lightweight, it is an ideal choice for this scenario as it can benefit from the instructor model knowledge while maintaining its efficiency. The depthwise separable convolutions and reduced parameter count in MobileNet allow it to learn compact representations, making it highly efficient when deployed on mobile devices or in edge computing environments. Furthermore, by using MobileNet as a student model, the KD process can help the model generalize better and achieve higher accuracy than training from scratch, all while maintaining low computational requirements. This makes MobileNet a suitable choice for applications that need to balance performance with real-time, low-latency execution. Fig 3 presents the architecture overview of the MobileNet model.
\begin{figure}[h!]
    \centering
    \includegraphics[width = 9cm]{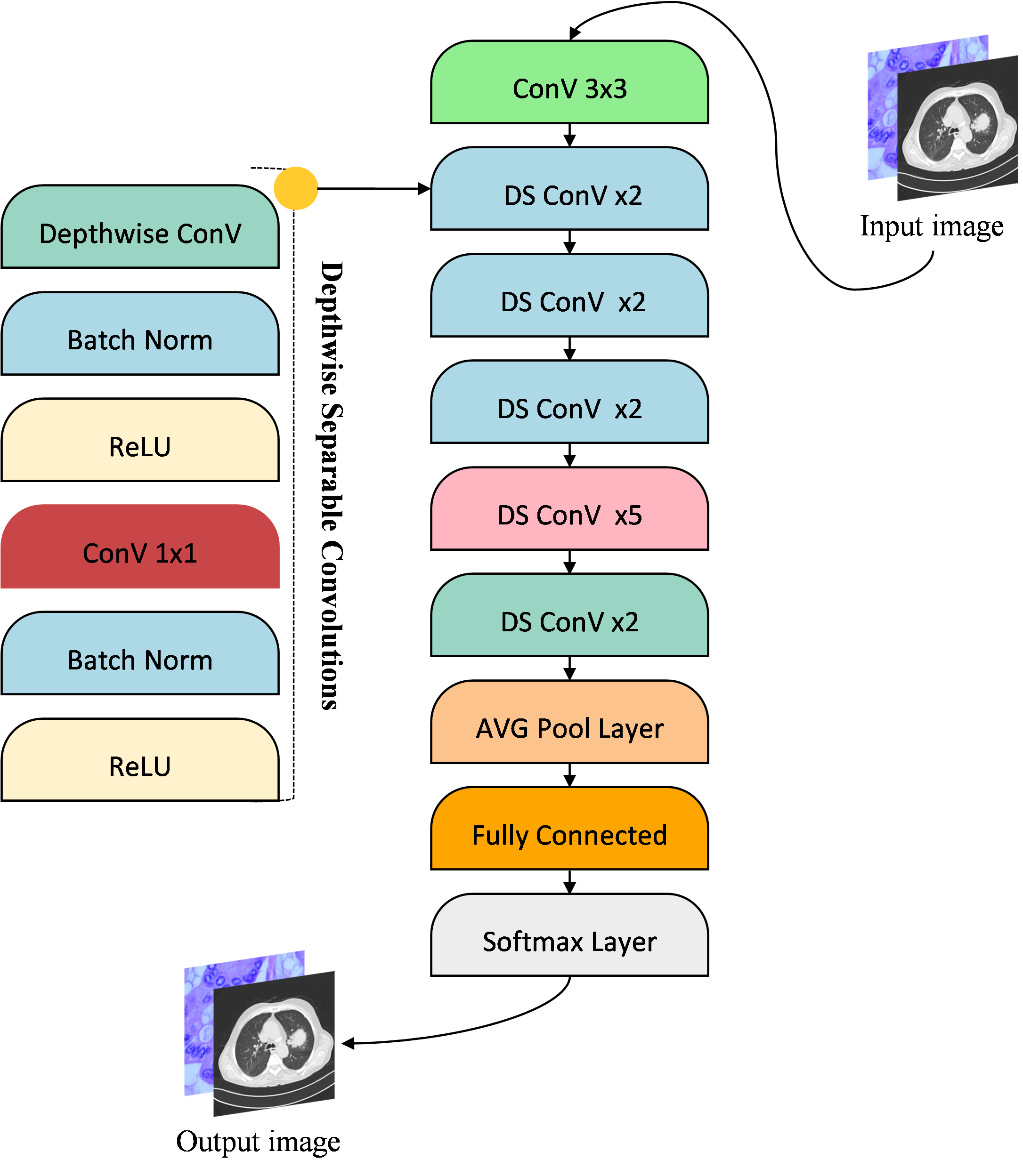}
    \caption{Architecture overview of selected student model}
    \label{fig:se.png}
\end{figure}
\subsection{Limitations of traditional KD Losses}
In conventional KD, the soft labels are calculated using a constant weight $\omega$ (0.1) for the KD loss. Although this method is often effective in many situations, it has significant drawbacks when the data has different levels of uncertainty. The degree of uncertainty in various areas of an image can vary greatly, especially in disease image analysis. For instance, certain image portions of a disease image can have easily identifiable, distinct features, while other portions might be less structured, noisy, or ambiguous, necessitating a more nuanced interpretation. The distillation method treats every region equally, regardless of the fluctuating uncertainty in the disease region, by employing a set weight $\omega$ (0.1). This method is unable to adjust to the complexity of disease images, where particular portions may have a larger degree of uncertainty than others, making it challenging to interpret noisy or confusing areas with accuracy. This could result in lower performance and decreased accuracy, especially in regions that need more attentive learning, as the model might not pay enough attention to places with significant uncertainty.

When calculating the soft labels with a fixed weight $\omega$ (0.1), the conventional KD loss is commonly represented as follows:

\begin{equation}
L_{KD} = \omega \times L_{\text{crossentropy}} (C - \hat{C}) + (1 - \omega) \times X^2 \times \text{Kullback-Leibler} (\text{soft}_{\phi}(X), \text{hard}_{\phi}) \tag{13}
\end{equation}
\begin{itemize}
    \item $L_{\text{crossentropy}}(C, \hat{C})$ is the difference in cross-entropy between the predicted labels ($\hat{C}$) and the genuine labels ($C$).
    \item The probability distribution is softened by the temperature parameter $X$.
    \item $ \text{soft}_\varphi(X), \text{hard}_\varphi $ is the difference between the hard target labels $\text{hard}_\varphi$ and the softened instructor model probability distribution $\text{soft}_\varphi(X)$ according to the Kullback-Leibler divergence (KL).
    \item With $\omega \in [0, 1]$, $\omega$ is the fixed weight that is used to balance the two loss terms.
\end{itemize}
The student model can learn from both the actual labels and the soft labels generated by the instructor model as follows in Equation (EQ-13), which combines the usual cross-entropy loss and the KD loss. The instructor model predictions are softened by the temperature \( T \). Each loss term's contribution to the overall loss function is controlled by the fixed weight \( \omega \).
\subsection{Optimizing Transformer KD Loss with Fuzzy Scaled Weights}
A constant weight $\omega$ (0.1)) that disregards the various levels of uncertainty in disease images limits the KD loss capacity to adapt to complex or ambiguous regions. We suggest using fuzzy weighting to scale the distillation loss and dynamically modify the soft label contribution according to sample uncertainty in order to improve the distillation process. Traditional methods use a fixed weight to transfer knowledge from the instructor model (VITB32) to the student model (MobileNet), ignoring the variable uncertainty or ambiguity present in disease images. We can modify the weight given to the KD loss during the knowledge transfer procedure by employing fuzzy logic.

\subsection*{Scaled Weights with Fuzzy Rules}
Instead of treating complete images consistently, the dynamic weighting modification utilizing fuzzy logic is intended to address the different levels of uncertainty within each disease image. This indicates that areas of high confidence and areas of ambiguity within the same image are implicitly recognized by the model. The fuzzy logic method dynamically assigns different weights during training, rather than explicitly separating the image into distinct zones. This allows the model to down-weight uncertain regions that could complicate learning and concentrate more on dependable (high-confidence) parts. The goal is to dynamically modify the student model reliance on the instructor instruction based on the local uncertainty in each image, rather than training numerous student model for distinct images. The usage of fuzzy logic membership functions is what causes the overlap between the low, medium, and high confidence levels. Fuzzy logic permits a single confidence score, say 0.5, to partially belong to numerous categories at once, in contrast to hard classifications where each value strictly belongs to one category. This implies that, to varying degrees, a confidence level of 0.5 can be regarded as medium and high. In particular, at 0.5, the confidence is at the point where medium membership (which spans 0.2 to 0.8) is strong, low membership ends, and high membership starts. In order to facilitate more flexible and seamless decision-making in uncertain situations, 0.5 is understood as overlapping between medium and high confidence levels rather than solely belonging to one. The same is true for the Uncertainty.
   
   As an alternative of a straightforward scaling factor, the output is a weight. This weight regulates the degree to which the KD loss affects training. By using a weighted method, the model can dynamically modify the instructor guidance contribution, assigning varying degrees of priority to the KD loss based on the level of confidence or uncertainty at certain phases or disease locations.
   \section*{Step-by-Step Overview of Fuzzy Rule: Scaling with Fuzzy Weighting}

\subsection*{Input Variables}

\textbf{Confidence}: This variable shows how confident the instructing model is in its estimates. It has membership functions for three levels and is defined between 0 and 1.

\[
\text{If ConfidenceValue} \geq 0 \text{ and ConfidenceValue} \leq 0.5:
\]
\[
\text{Confidence} = \text{"Low"} \quad \text{Return Confidence}
\]
\[
\text{Else if ConfidenceValue} > 0.2 \text{ and ConfidenceValue} \leq 0.8:  
\]
\[
\text{Confidence} = \text{"Medium"} \quad \text{Return Confidence}
\]
\[
\text{Else if ConfidenceValue} > 0.5 \text{ and ConfidenceValue} \leq 1:  
\]
\[
\text{Confidence} = \text{"High"} \quad \text{Return Confidence}
\]
\textbf{Uncertainty}: The degree of ambiguity in the medical image under analysis is captured by this variable. It has three levels and is specified in the range of 0 to 1.

\[
\text{If UncertaintyValue} \geq 0 \text{ and UncertaintyValue} \leq 0.4:
\]
\[
\text{Uncertainty} = \text{"Low"} \quad \text{Return Uncertainty}
\]
\[
\text{Else if UncertaintyValue} > 0.3 \text{ and UncertaintyValue} \leq 0.9:  
\]
\[
\text{Uncertainty} = \text{"Medium"} \quad \text{Return Uncertainty}
\]
\[
\text{Else if UncertaintyValue} > 0.7 \text{ and UncertaintyValue} \leq 1:  
\]
\[
\text{Uncertainty} = \text{"High"} \quad \text{Return Uncertainty}
\]

\subsection*{Output Variable}

\textbf{Weight}: The weight given to the KD loss is determined by this variable. It has three levels and is defined between 0 and 1.

\[
\text{If WeightValue} \geq 0 \text{ and WeightValue} \leq 0.4:
\]
\[
\text{Weight} = \text{"Low"} \quad \text{Return Weight}
\]
\[
\text{Else if WeightValue} > 0.3 \text{ and WeightValue} \leq 0.7:  
\]
\[
\text{Weight} = \text{"Medium"} \quad \text{Return Weight}
\]
\[
\text{Else if WeightValue} > 0.6 \text{ and WeightValue} \leq 1:  
\]
\[
\text{Weight} = \text{"High"} \quad \text{Return Weight}
\]

\subsection*{Loss Calculation}

\[
L_{\text{total}} = v \times (Fuzzy_{\text{Weight}} \times L_{\text{KD}}) + (1 - v) \times L_{\text{CE}}
\]

Where:
\begin{itemize}
    \item $L_{\text{total}}$: The overall loss incurred during the student model training.
    \item $L_{\text{KD}}$: The KD loss quantifies the discrepancy between the outputs of the instructor and students.
    \item $L_{\text{CE}}$: Cross-Entropy classification loss between the true labels and the predictions of the student model.
    \item $v$: Hyperparameter to maintain a balance between the significance of classification loss and KD loss.
    \item $Fuzzy_{\text{Weight}}$: The fuzzy logic approach dynamic weight, which ranges from 0 to 1, adaptively modifies the KD loss strength based on uncertainty and confidence.
\end{itemize}

\subsection*{Scaling Rules}

\begin{itemize}
    \item \textbf{Rule 1}: The weight is low when uncertainty is high and confidence is low. This suggests that when there is a lot of uncertainty, the model should rely less on the instructor's weight prediction.
    \item \textbf{Rule 2}: The weight is medium if the levels of uncertainty and confidence are both medium. This indicates a reasonable dependence on the instructor's weight prediction.
    \item \textbf{Rule 3}: The weight is high if there is a high degree of confidence and a low level of uncertainty. This implies that when the instructor is certain and the data is not ambiguous, the model should heavily rely on the instructor.
\end{itemize}

\subsection*{Result}

The rules are used to establish a control system, and a Scaled Weight System Simulation is configured to calculate the output weight depending on the input values for uncertainty and confidence. 

The fuzzy logic system uses overlapping membership functions to handle situations where both the confidence and uncertainty values are close to the boundaries of their designated levels, permitting partial membership in adjacent levels. This indicates that these memberships are combined and fuzzy inference (FI) procedures are applied to get the output weight. For instance. Confidence = 0.3 → may concurrently fall into the Low and Medium ranges (to varying degrees). Uncertainty = 0.7 → may concurrently fall into the Medium and High ranges (Algorithm 1). 
\begin{algorithm}
\caption{Determining Output Weight Level Based on Confidence and Uncertainty}
\begin{algorithmic}[1]
\State \textbf{Input:} Confidence level (Low, Medium), Uncertainty level (Medium, High)
\State \textbf{Output:} Weight Level

\If{ConfidenceLevel == Low and UncertaintyLevel == Medium}
    \State WeightLevel = "Lower Weight"
    \State \textbf{Return} WeightLevel
\ElsIf{ConfidenceLevel == Low and UncertaintyLevel == High}
    \State WeightLevel = "Lower Weight"
    \State \textbf{Return} WeightLevel
\ElsIf{ConfidenceLevel == Medium and UncertaintyLevel == Medium}
    \State WeightLevel = "Moderate Weight"
    \State \textbf{Return} WeightLevel
\ElsIf{ConfidenceLevel == Medium and UncertaintyLevel == High}
    \State WeightLevel = "Moderate to Lower Weight"
    \State \textbf{Return} WeightLevel
\EndIf
\end{algorithmic}
\end{algorithm}
\subsection*{Fuzzy Memberships}
The output weight, which reflects the combined effect of moderate confidence and moderate-to-high uncertainty, would fall between low and medium due to overlapping fuzzy memberships. The precise amount varies according to the rule definitions and membership levels, but in general, it promotes moderate reliance on KD loss when confidence is medium and less dependence when uncertainty is high. These two metrics are completely dependent on one another and totally complementary if uncertainty is simply defined as Uncertainty = (1-Confidence). If so, adding both independently to fuzzy rules creates duplication. This suggests that since uncertainty does not give extra independent information, the fuzzy scale weight system output weight could be efficiently decided only by the confidence level. After that, the weight assignment procedures become a straightforward or imprecise mapping between output weight and confidence: Weight increases with high confidence, weight decreases with medium confidence, and weight decreases with low confidence. This dynamic scaling technique improves the distillation process and yields better results, particularly in complex disease analysis, by allowing the model to focus on areas with varying degrees of certainty.

\subsection*{Confidence Membership Functions}
\textbf{Set\_Low (confidence):}
\[
\text{Set\_Low (confidence)} =
\begin{cases}
1 & \text{if } \text{confidence} \leq 0.2 \\
\frac{0.5 - \text{confidence}}{0.3} & \text{if } 0.2 < \text{confidence} \leq 0.5 \\
0 & \text{if } \text{confidence} > 0.5
\end{cases}
\]

\textbf{Set\_Medium (confidence):}
\[
\text{Set\_Medium (confidence)} =
\begin{cases}
0 & \text{if } \text{confidence} \leq 0.2 \\
\frac{\text{confidence} - 0.2}{0.3} & \text{if } 0.2 < \text{confidence} \leq 0.5 \\
\frac{0.8 - \text{confidence}}{0.3} & \text{if } 0.5 < \text{confidence} \leq 0.8 \\
0 & \text{if } \text{confidence} > 0.8
\end{cases}
\]

\textbf{Set\_High (confidence):}
\[
\text{Set\_High (confidence)} =
\begin{cases}
0 & \text{if } \text{confidence} < 0.5 \\
\frac{\text{confidence} - 0.5}{0.5} & \text{if } 0.5 \leq \text{confidence} \leq 1
\end{cases}
\]

\subsection*{Uncertainty Membership Functions}
\textbf{Set\_Low (uncertainty):}
\[
\text{Set\_Low (uncertainty)} =
\begin{cases}
1 & \text{if } \text{uncertainty} \leq 0.2 \\
\frac{0.4 - \text{uncertainty}}{0.2} & \text{if } 0.2 < \text{uncertainty} \leq 0.4 \\
0 & \text{if } \text{uncertainty} > 0.4
\end{cases}
\]

\textbf{Set\_Medium (uncertainty):}
\[
\text{Set\_Medium (uncertainty)} =
\begin{cases}
0 & \text{if } \text{uncertainty} \leq 0.3 \\
\frac{\text{uncertainty} - 0.3}{0.3} & \text{if } 0.3 < \text{uncertainty} \leq 0.6 \\
\frac{0.9 - \text{uncertainty}}{0.3} & \text{if } 0.6 < \text{uncertainty} \leq 0.9 \\
0 & \text{if } \text{uncertainty} > 0.9
\end{cases}
\]

\textbf{Set\_High (uncertainty):}
\[
\text{Set\_High (uncertainty)} =
\begin{cases}
0 & \text{if } \text{uncertainty} < 0.7 \\
\frac{\text{uncertainty} - 0.7}{0.3} & \text{if } 0.7 \leq \text{uncertainty} \leq 1
\end{cases}
\]
\subsection{Overview of the proposed FuzzyDistillViT-MobileNet: Scaled Weights for KD Loss }
The proposed FuzzyDistillViT-MobileNet approach integrates fuzzy logic to scale the weights for KD loss, aiming to enhance model performance in complex disease image classification tasks, such as the diagnosis of LC. In traditional KD, the soft labels generated by the instructor model are transferred to the student model using a fixed weight, typically $\omega$ (0.1). However, this constant weight approach does not consider the varying degrees of uncertainty present in different regions of a disease image, which can affect model learning, particularly when some portions of the image are more ambiguous or noisy. The FuzzyDistillViT-MobileNet addresses this limitation by dynamically adjusting the weight assigned to the KD loss based on the uncertainty and confidence levels of different regions in the image. Fuzzy logic is employed to adaptively scale the KD loss weight, allowing the student model to focus more on confident regions while reducing reliance on uncertain areas. This is achieved through fuzzy membership functions, which categorize the confidence and uncertainty levels into low, medium, and high ranges. By using fuzzy rules, the weight applied to the KD loss is adjusted according to the local uncertainty and confidence at each stage of training. The dynamic scaling enables the model to allocate more attention to high-confidence areas, ensuring that regions with higher uncertainty or ambiguity do not unduly influence the learning process. The ViTB32 model serves as the instructor model in this setup, guiding the MobileNet student model through knowledge transfer. ViTB32, with its self-attention mechanisms, excels at capturing global context in images, which is essential for recognizing fine-grained details in medical images. MobileNet, known for its computational efficiency, is chosen as the student model to ensure that the model remains lightweight, making it suitable for real-time, resource-constrained applications. The KD process between ViT and MobileNet, augmented by the fuzzy logic-based weight scaling, ensures that the student model not only learns efficiently but also generalizes better to the varying uncertainty levels in the images, ultimately improving classification accuracy for LC detection. Through this approach, the FuzzyDistillViT-MobileNet framework effectively addresses the challenges associated with uncertainty in LC image analysis, providing a more robust solution for classifying complex disease images. By dynamically adjusting the weight of the KD loss, the model is able to learn in a way that emphasizes high-confidence features while reducing the impact of ambiguous regions, resulting in improved model performance and reliability. This method is particularly beneficial for disease diagnoses, where precision and attention to uncertain regions are critical for accurate outcomes. Fig 4 presents the overview of the proposed FuzzyDistillVit-MobileNet framework.
\begin{figure}[h!]
    \centering
    \includegraphics[width = 12cm]{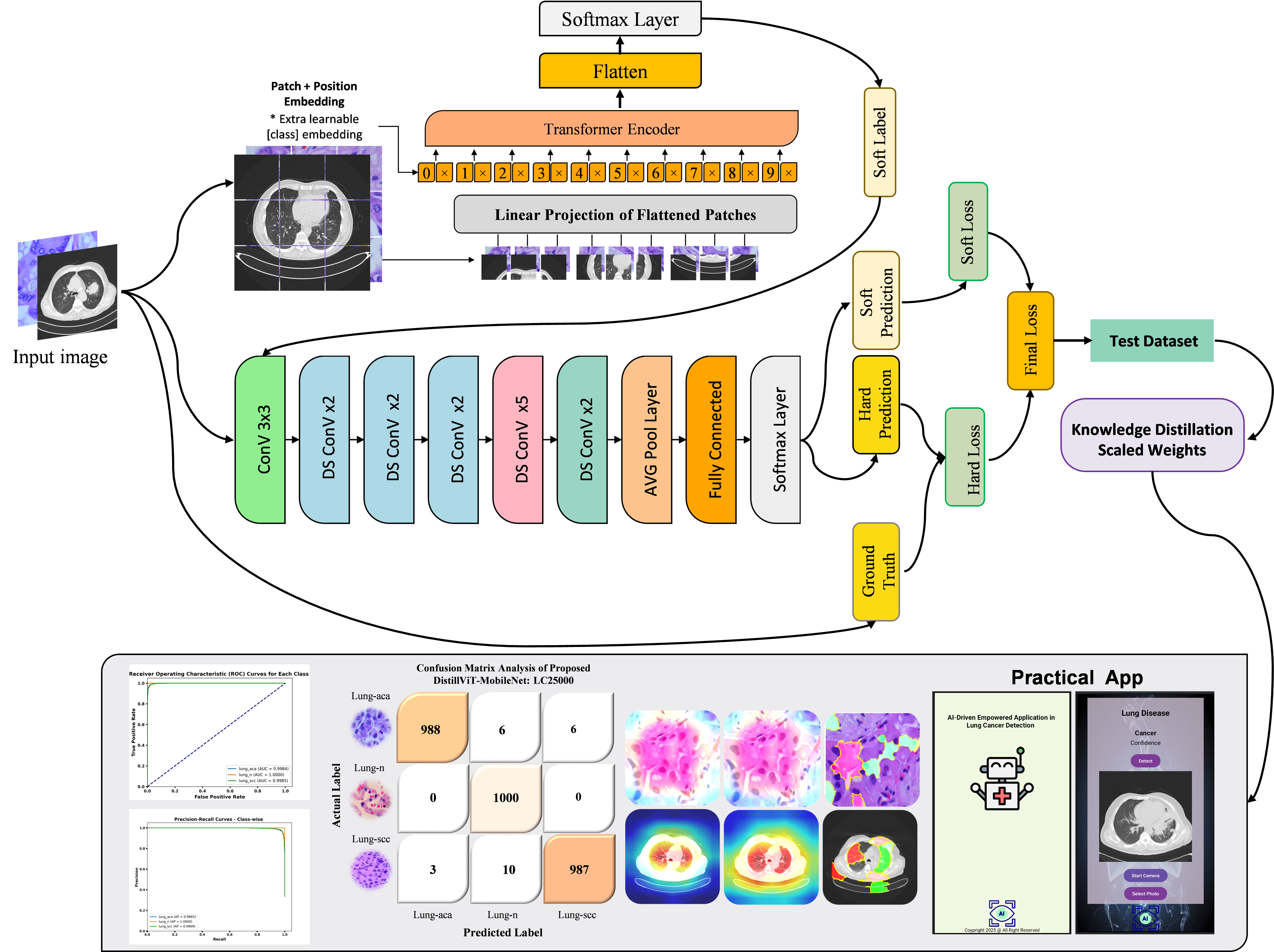}
    \caption{Overview of the proposed FuzzyDistillVit-MobileNet framework}
    \label{fig:se.png}
\end{figure}
\subsection{Experimental setting and hyperparameter details}
We carefully chose a set of hyperparameters to guarantee reliable and efficient model training. To attain adequate learning without overfitting, the model was trained for 30 epochs. To maximize convergence, a learning rate of 0.001 was employed. To strike a balance between computational efficiency and model fidelity, a batch size of 64 was used. Because of its quick convergence and capacity to modify learning rates for specific parameters, the Adam optimizer was chosen to stabilize training and enhance performance. The model learning process was efficiently guided by the categorical cross-entropy loss function, which made it ideal for multi-class lung disease categorization. A 70:10:20 ratio was used to divide the dataset into training, validation, and test sets in order to prevent overfitting and ensure a robust model evaluation. Twenty percent of the data was retained as a test set for assessing the model performance. Ten percent of the data was utilized as a validation set to track model generalization, while the remaining eighty percent was used for the training phase. By providing a precise evaluation of the model performance in actual situations, this partitioning technique guarantees that the model is tested on unseen data. The distribution of the LC dataset is shown in Table 2.
\subsection{Evaluation metrics for FuzzyDistillViT-MobileNet model}
We used well-known assessment indicators to do a comprehensive evaluation of the FuzzyDistillViT-MobileNet model performance. The results aligned with previous research findings, confirming the validity of our technique. For each sample in the lung disease diagnosis dataset, the model predictions were categorized using four primary categories: True Positive (TP), True Negative (TN), False Positive (FP), and False Negative (FN). Commonly used metrics like accuracy (EQ-14), precision (EQ-15), recall (EQ-16), and F1-score (EQ-17) which offer numerical insight into how effectively the model distinguishes between correct and incorrect classifications were utilized to evaluate the performance.
\begin{equation}
\text{Accuracy} = \frac{TP + TN}{TP + FP + TN + FN} \tag{14}
\end{equation}

\begin{equation}
\text{Precision} = \frac{TP}{TP + FP} \tag{15}
\end{equation}

\begin{equation}
\text{Recall} = \frac{TP}{TP + FN} \tag{16}
\end{equation}

\begin{equation}
\text{F1-Score} = \frac{2 \times (\text{Precision} \times \text{Recall})}{\text{Precision} + \text{Recall}} \tag{17}
\end{equation}

\section{Results and implementation}\label{sec13}
In this section. Section 4.1 explores pixel-level enhancement strategies to refine image quality, followed by dataset preprocessing and balancing techniques in Section 4.2 to ensure uniformity and consistency across the data. The performance of the FuzzyDistillViT-MobileNet model is thoroughly evaluated in Sections 4.3 through 4.6, where the impact of various datasets, such as the LC25000 histogram and IQOTH/NCCD CT-scan, is analyzed in comparison to state-of-the-art (SOTA) CNN models. Section 4.7 delves into the interpretability and visualization of the model's decisions, employing methods like GRADCAM, GRADCAM++, and LIME to provide transparent insights into model behavior. Section 4.8 includes additional testing to further assess model robustness, while Section 4.9 discusses real-time application testing on Android devices, showcasing the model's practical deployment. Finally, Section 4.10 presents an ablation study to evaluate the contributions of various components of the proposed method, ensuring a comprehensive understanding of its effectiveness and performance.
\subsection{Optimizing Image Quality Through Pixel-Level enhancement}
This section provides a thorough overview of sample enhancement using DWT fusion, including techniques for enhancing image quality such as Gamma Correction and Histogram Equalization. These techniques aim to increase image qualities by using advanced enhancement techniques and pixel-level fusion.
\subsubsection{Pixel-Level 1\textsuperscript{st} enhancement}
By applying a power-law transformation to pixel intensities, \textit{gamma image enhancement}, also referred to as \textit{gamma correction}, is a non-linear technique that modifies the brightness and contrast of digital images. Using the formula\[\text{GAMMA}_{\text{out}} = M_A \cdot \text{GAMMA}_{\text{in}}^{\phi}\] the procedure uses \(\text{GAMMA}_{\text{in}}\) and \(\text{GAMMA}_{\text{out}}\) to represent input and output pixel values, \(M_A\) as a scaling constant, and \(\phi\) (gamma) to determine the type of adjustment. For underexposure correction, a gamma value less than one brightens the image by shifting mid-tones toward higher intensities; for overexposed scenes, a gamma value greater than one darkens the image by shifting mid-tones downward. By adjusting specific mid-tone areas, gamma correction, in contrast to linear techniques, maintains details in highlights and shadows.
\subsubsection{Pixel-Level 2\textsuperscript{nd} enhancement}
Histogram Equalization improves image contrast by redistributing pixel intensities across the entire image, stretching the histogram to cover the full intensity range. It works by calculating the cumulative distribution function (CDF) of the image intensity values and then mapping each pixel to a new intensity based on this function. This process equalizes the image histogram, making the overall brightness and contrast more uniform. However, it can sometimes lead to over-enhancement in areas with high contrast, amplifying noise. The image is processed as a whole, unlike techniques like CLAHE, which operate locally. The transformation is typically computed globally across the entire image, with pixel intensities mapped according to their cumulative frequency. 
\subsubsection{Fusion Pixel 1\textsuperscript{st} with Pixel 2\textsuperscript{nd} Level by Level}
Two processed inputs (Pix1 and Pix2) are combined using wavelet-based techniques in the Pixel 1st with Pixel 2nd Level by Level fusion method shown in Fig 5. The wavedec2 function is used to first standardize both images to a 224x224 resolution and then break them down into multi-scale frequency components. A mean operation is used to fuse the coefficients at each decomposition level. This method is performed recursively to succeeding levels, averaging the coefficients at the first level. The fused image is then scaled to the 0-255 intensity range for consistency after the combined coefficients have undergone inverse wavelet modification. Through the integration of complementary spatial and frequency information from both inputs, this method produces a combined output that is enhanced for further tasks such as feature extraction or analysis (Algorithm-2).
\begin{figure}[h!]
    \centering
    \includegraphics[width = 9cm]{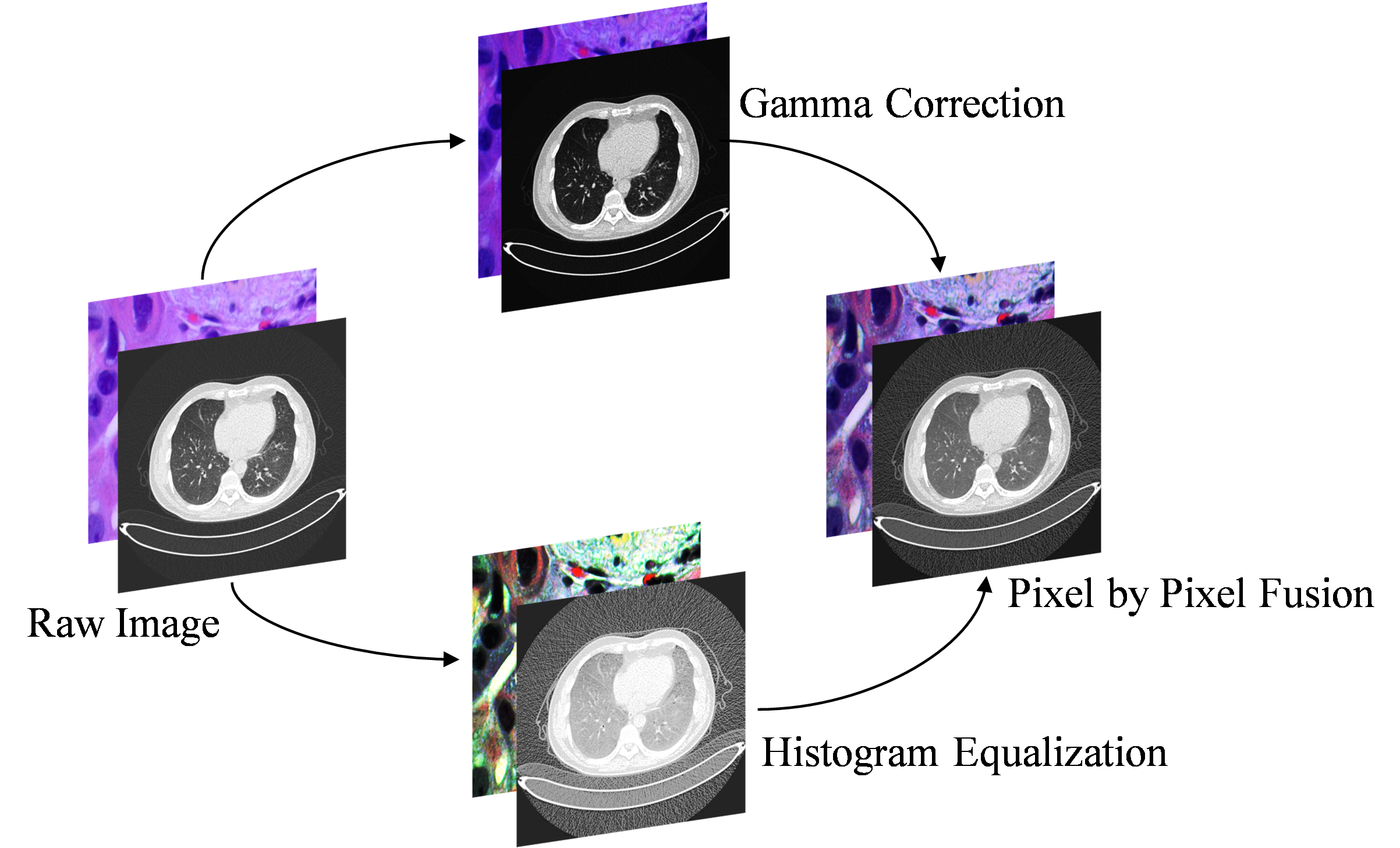}
    \caption{Techniques for optimizing image quality through Pixel-Level enhancement}
    \label{fig:se.png}
\end{figure}
\begin{algorithm}
\caption{Image Fusion Algorithm}
\begin{algorithmic}[1]
\State \textbf{Input:} Two Process Images → $\rho_1$ (Pixel-Level 1st) and $\rho_2$ (Pixel-Level 2nd)
\State \textbf{Initialization:} Set Fusion Method Options → Mean (Pn)
\State \textbf{Load Images:} $\rho_1$ and $\rho_2$ \textbf{(Ensure the images are the same size: 224x224)}
\State \textbf{Pixel-level Image Fusion:} 
\State \quad $\text{Coef}_1 \gets \text{wavedec2} (I_1, \text{wavelet})$
\State \quad $\text{Coef}_2 \gets \text{wavedec2} (I_2, \text{wavelet})$
\State \textbf{Coefficients fused:} Initialize blank list for fused coefficients $\left( \text{fusedCoef} \gets [] \right)$
\For{each level image $i$ in the wavelet coefficients (excluding the final one)}
    \If{$i = 0$ (Pixel-Level 1st)}
        \State \quad FUSION\_METHOD = mean
        \State \quad $\text{Coefficientsfused}[i] \gets \text{Coefficientsfused} (\text{Coef1}[0], \text{Coef2}[0], \text{FUSION\_METHOD})$
    \Else \text{(upcoming levels)}
        \State \quad Pix1 $\gets \text{Coefficientsfused} (\text{Coef1}[i][0], \text{Coef2}[i][0], \text{FUSION\_METHOD})$
        \State \quad Pix2 $\gets \text{Coefficientsfused} (\text{Coef1}[i][1], \text{Coef2}[i][1], \text{FUSION\_METHOD})$
        \State \quad Pix3 $\gets \text{Coefficientsfused} (\text{Coef1}[i][2], \text{Coef2}[i][2], \text{FUSION\_METHOD})$
        \State \quad Append (Pix1, Pix2, Pix3) to \text{Coefficientsfused}
    \EndIf
\EndFor
\State \textbf{Fused Image Redraw:}
\State \quad $\text{fusedPix} \gets \text{wavedec2} (\text{Coefficientsfused}, \text{wavelet})$
\State \textbf{Adjust the fused image's pixel values to fall between 0 and 255.}
\State \quad $\text{fusedImage} \gets \text{normalize} (\text{fusedImage})$
\State \textbf{Save} $\text{fusedPix} (i)$
\end{algorithmic}
\end{algorithm}
\subsection{Dataset preprocessing and balancing techniques}
This section offers a thorough detail of the datasets utilized in the research, together with information on the number of classes and samples in each dataset. We outline the preparation methods, such as scaling and normalization, that were used on the datasets. We also used data augmentation techniques, which artificially increase the diversity of the training data to improve model generalization and balance the dataset. Fusion Pixel 1st with Pixel 2nd Level by Level, next steps involve preprocessing and augmentation to prepare the data for model training.

Preprocessing and augmentation involve resizing the raw images to 224 by 224 pixels. With this scaling, the input dimensions for deep learning (DL) models are guaranteed to be uniform and standardized. After that, the photos are rescaled by normalizing the pixel values to fall between [0, 1]. In this stage, the data range is uniformized to avoid problems with high or tiny values that can slow down convergence, ensuring that the model trains efficiently.

Data augmentation methods are used after preprocessing to improve models' capacity for good generalization. These methods include flips in the horizontal and vertical directions, as well as random rotations (90°). Because each of these methods creates new variations of the original photographs, the dataset becomes more diverse. While flips increase the model's resilience to variations in the spatial arrangement of features, random rotations assist the model in becoming invariant to orientation. Through the efficient expansion of the training dataset with changes that the model may meet in real-world scenarios, this augmentation procedure is essential for enhancing the robustness and performance of models. Fig. 6 shows the preparation pipeline as a whole.
\begin{figure}[h!]
    \centering
    \includegraphics[width = 10cm]{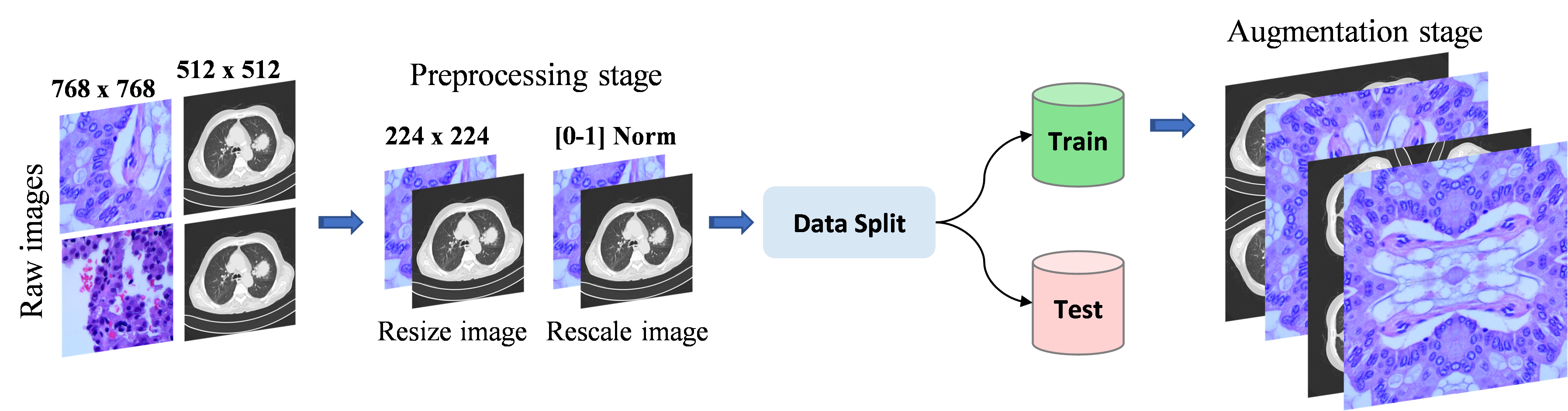}
    \caption{Techniques for preprocessing and augmentation to improve the diversity of the dataset and standardize it}
    \label{fig:se.png}
\end{figure}
Table 2 presents the dataset distribution before and after preprocessing for two utilize datasets: IQOTH/NCCD and LC25000. For the IQOTH/NCCD dataset, the classes include Benign, Malignant, and Normal. Before preprocessing, the training set contains 86 Benign, 339 Malignant, and 332 Normal images, with validation sets having 10 Benign, 49 Malignant, and 33 Normal images. After preprocessing, the training set for each class has been standardized to 339 images, and the validation set has 49 images for each class, with the test set containing 24 Benign, 113 Malignant, and 84 Normal images. In the LC25000 dataset, which includes Lung-aca, Lung-n, and Lung-scc classes, each class initially had 3600 training samples and 400 validation samples, with 1000 samples for testing. After preprocessing, the number of images remains consistent across the training and validation sets for each class, maintaining 3600 training, 400 validation, and 1000 test images per class.
\begin{table}[h!]
\centering
\caption{Dataset Distribution before and after balancing LC25000 and IQOTH/NCCD datasets}
\begin{tabular}{llccccc}
\hline
\textbf{Dataset} & \textbf{Class} & \textbf{Before Train} & \textbf{Before Valid} & \textbf{After Train} & \textbf{After Valid} & \textbf{Test} \\
\hline
\multirow{3}{*}{IQOTH/NCCD} & Benign & 86 & 10 & 339 & 49 & 24 \\
 & Malignant & 339 & 49 & 339 & 49 & 113 \\
 & Normal & 332 & 33 & 339 & 49 & 84 \\
\multirow{3}{*}{LC25000} & Lung-aca & 3600 & 400 & 3600 & 400 & 1000 \\
 & Lung-n & 3600 & 400 & 3600 & 400 & 1000 \\
 & Lung-scc & 3600 & 400 & 3600 & 400 & 1000 \\
\hline
\end{tabular}
\end{table}
\subsection{Performance evaluation of the FuzzyDistillViT-MobileNet: Each Class}
Table 3 presents performance metrics (Precision, Recall, F1-Score, and Accuracy) for different classes across two datasets: Lung-Histopathology (LC25000) and IQ-OTHNCCD. In the Lung-Histopathology dataset, the Lung-aca class achieves the highest accuracy of 99.16\%, followed closely by Lung-scc with an accuracy of 99.36\%. The Lung-n class exhibits the lowest F1-score of 97.25\%, indicating relatively weaker performance compared to the other two Lung classes. In the IQ-OTHNCCD dataset, the Benign class leads in F1-Score (99.88\%) and achieves the highest accuracy of 99.54\%, while the Malignant class has a slightly lower F1-score of 98.28\% and accuracy of 99.12\%. The Normal class shows a strong performance with an F1-Score of 99.45\% and accuracy of 99.54\%, similar to the Benign class, but slightly lower in recall and F1-score. Overall, the results suggest that the Lung-scc and IQ-OTHNCCD Normal and Benign classes perform the best across most metrics, with Lung-n and Malignant showing slightly reduced performance.
\begin{table}[h!]
\centering
\caption{Classwise performance analysis of proposed FuzzyDistillViT-MobileNet on the LC25000 and IQOTH/NCCD dataset}
\begin{tabular}{cccccc}
\hline
\textbf{Datasets} & \textbf{Class} & \textbf{Precision} & \textbf{Recall} & \textbf{F1-Score} & \textbf{Accuracy} \\
\hline
\multirow{3}{*}{Lung-Histopathology (LC25000)} & Lung-aca & 98.44 & 98.30 & 99.12 & 99.16\% \\
 & Lung-scc & 99.87 & 99.75 & 99.36 &  \\
 & Lung-n & 99.12 & 98.35 & 97.25 &  \\
\multirow{3}{*}{IQ-OTHNCCD} & Benign & 99.56 & 99.48 & 99.88 & 99.54\% \\
 & Malignant & 99.12 & 98.10 & 98.28 &  \\
 & Normal & 99.54 & 99.85 & 99.45 &  \\
\hline
\end{tabular}
\end{table}
The confusion matrix results for FuzzyDistillViT-MobileNet model, as shown in fig 7(a) and fig 7(b), provide insights into its classification performance. In fig 7(a), which represents the LC25000 dataset, the model excels in correctly identifying Lung-n and Lung-scc samples, with 1000 and 987 correct predictions, respectively. However, there are some misclassifications: Lung-aca samples are misclassified as Lung-n (6 times) and Lung-scc (6 times). Despite this, the model demonstrates strong accuracy for most classes, with Lung-n being perfectly predicted, indicating good differentiation between Lung-aca and Lung-n. In fig 7(b), representing the IQOTH/NCCD dataset, the model shows high accuracy in distinguishing Malignant samples, with 112 correctly predicted and only one misclassified as Normal. The Benign class, while mostly accurate, has some confusion with Malignant (24 misclassifications). The Normal class is predominantly well-predicted, with just 1 misclassification into the Malignant category. Overall, the model performs well in distinguishing Benign, Malignant, and Normal classes, with only slight misclassification occurring in the Benign and Normal categories, demonstrating effective classification across the dataset.
\begin{figure*}[h]
\centering
\begin{minipage}[]{6cm}
  \centering
  \includegraphics[width = 5cm]{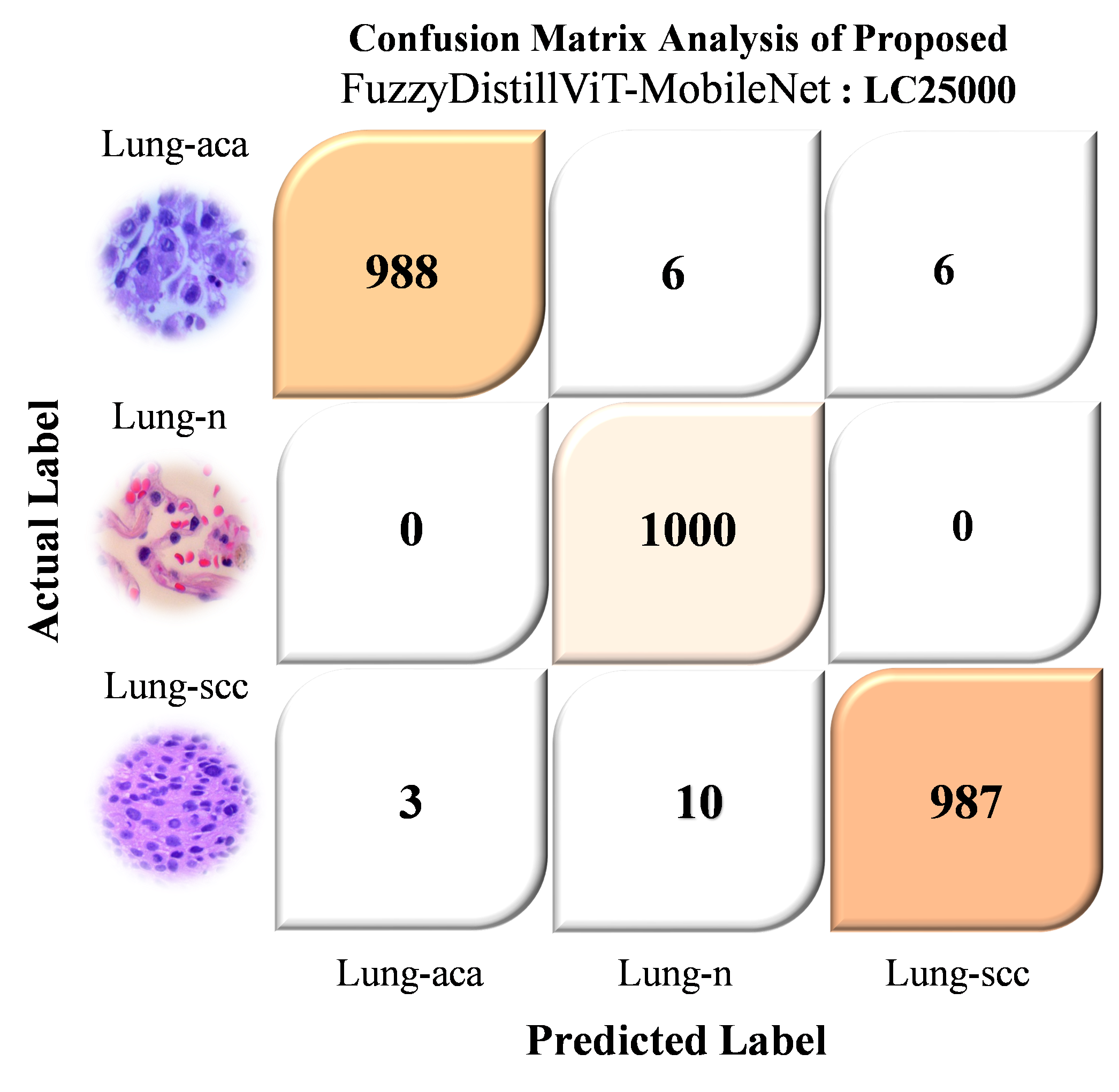}
  \vspace{-0.5cm} % Adds some space between the image and text
    \begin{center}
    \textbf{(a)}    
    \end{center}
\end{minipage}
\begin{minipage}[]{6cm}
  \centering
  \includegraphics[width = 5cm]{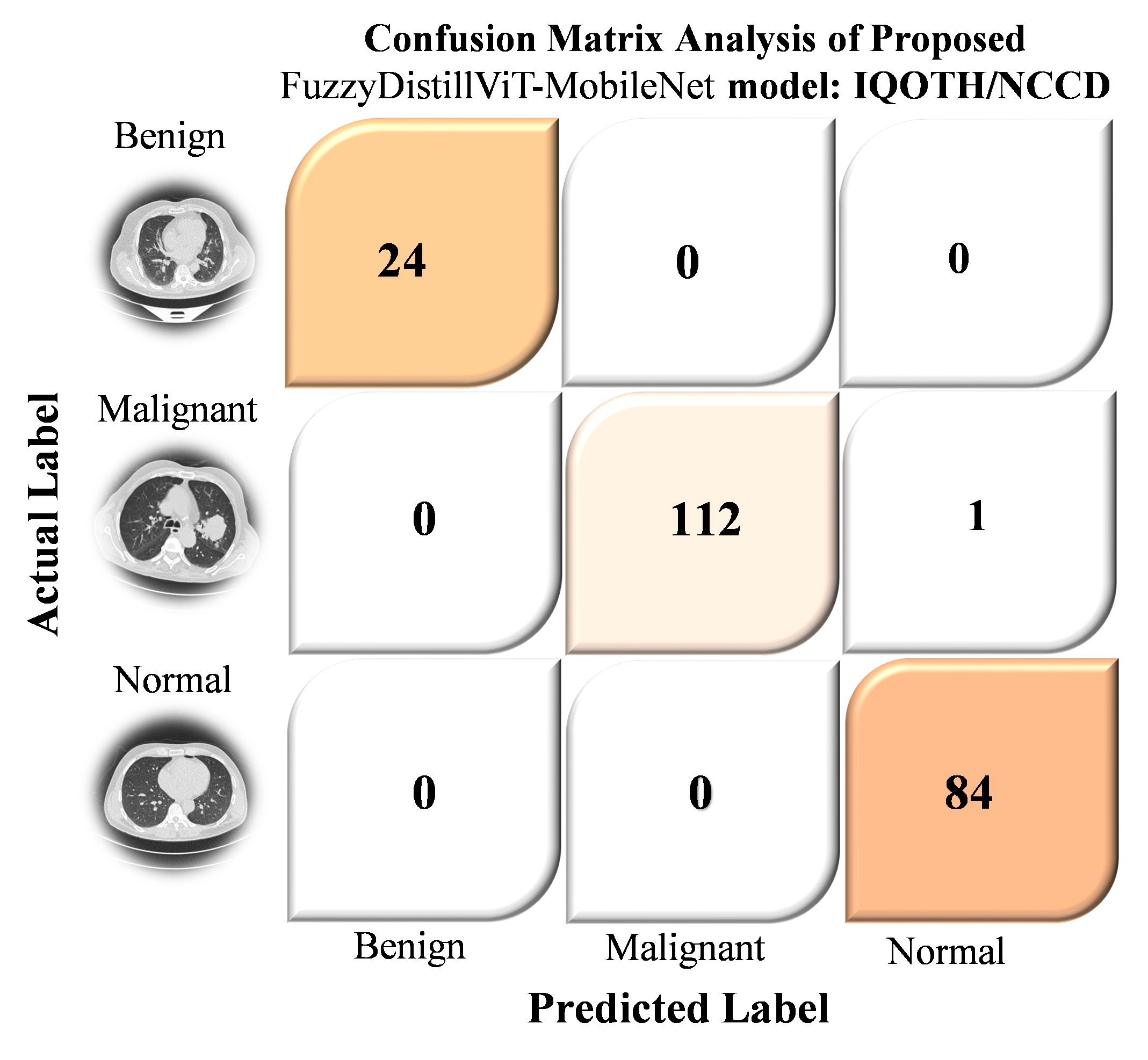}
  \vspace{-0.5cm} % Adds some space between the image and text
    \begin{center}
    \textbf{(b)}    
    \end{center}
\end{minipage}
  \caption{Visual investigation of confusion matrices for the proposed FuzzyDistillViT-MobileNet model using the (a) LC25000 and (b) IQ-OTHNCCD datasets}
  \label{fig:15.png}
\end{figure*}
The Precision-Recall (PR) and Receiver Operating Characteristic (ROC) curves for proposed FuzzyDistillViT-MobileNet model on the Histopathological LC25000 dataset show strong performance across the classes as presented in Fig 8. The PR curves in fig(a) demonstrate high precision and recall values, indicating that the model effectively identifies true positives for each class, particularly for lung-acc with an average precision (AP) score of 1.0000. Although lung-scc and lung-n also show efficient precision-recall curves, lung-aca leads with the highest AP, confirming its superior classification ability within this dataset. On the other hand, the ROC curves in fig(b) further support these findings by showing high True Positive Rates (TPR) for all classes and low False Positive Rates (FPR), particularly for lung-aca, which enhanced the highest AUC (Area Under the Curve) value of 0.9994. The AUC for the other classes is also impressive, with lung-scc and lung-n showing AUC values close to 1.0, reinforcing the model strong overall performance. In comparison, the PR curve focuses more on the precision-recall tradeoff, while the ROC curve emphasizes the model ability to distinguish between classes without being influenced by the class imbalance. Overall, the proposed model exhibits excellent classification, with particularly high performance in lung-aca, as seen in both PR and ROC analyses.
\begin{figure*}[h]
\centering
\begin{minipage}[]{6cm}
  \centering
  \includegraphics[width = 5cm]{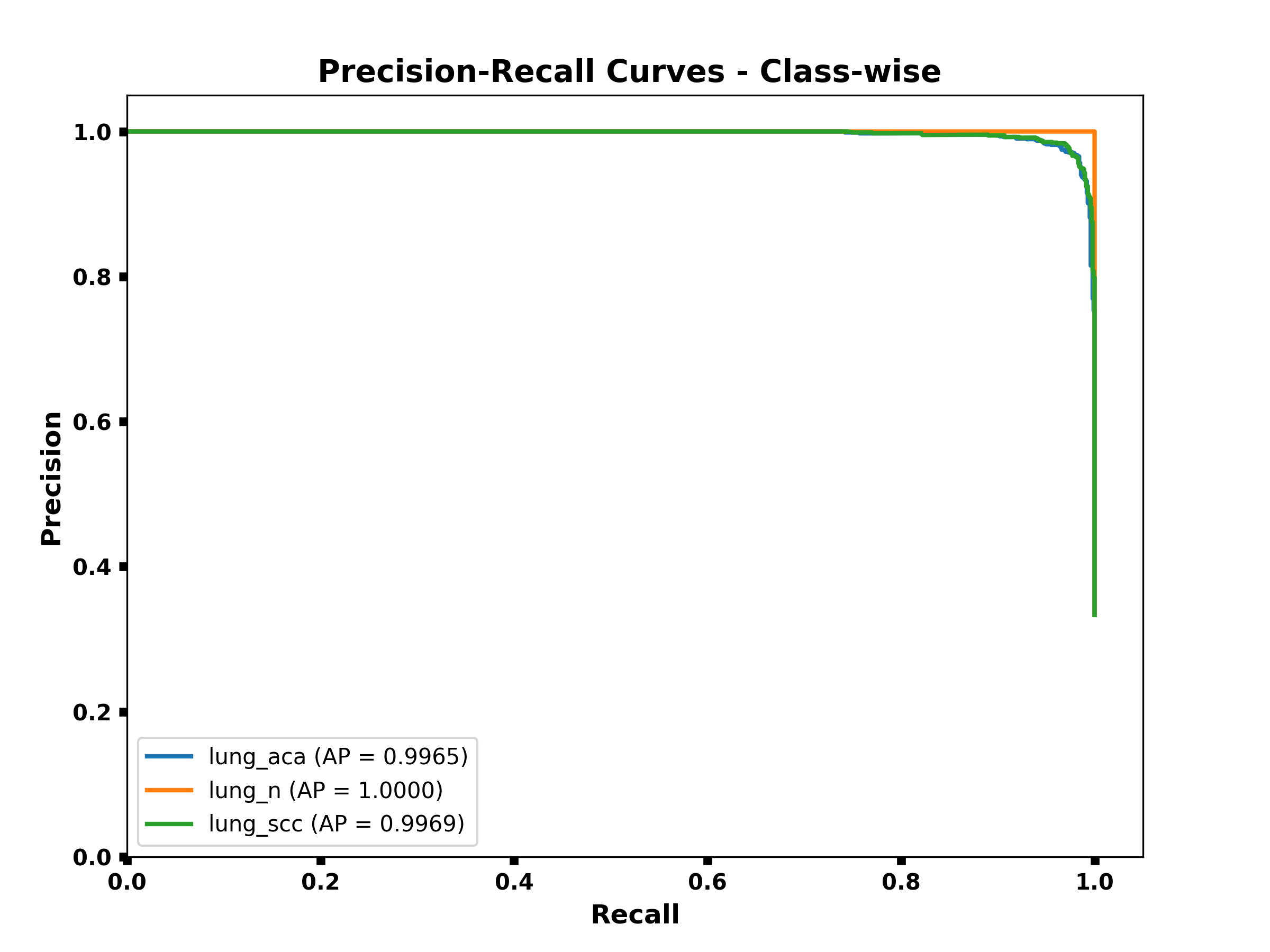}
  \vspace{-0.5cm} % Adds some space between the image and text
    \begin{center}
    \textbf{(a)}    
    \end{center}
\end{minipage}
\begin{minipage}[]{6cm}
  \centering
  \includegraphics[width = 5cm]{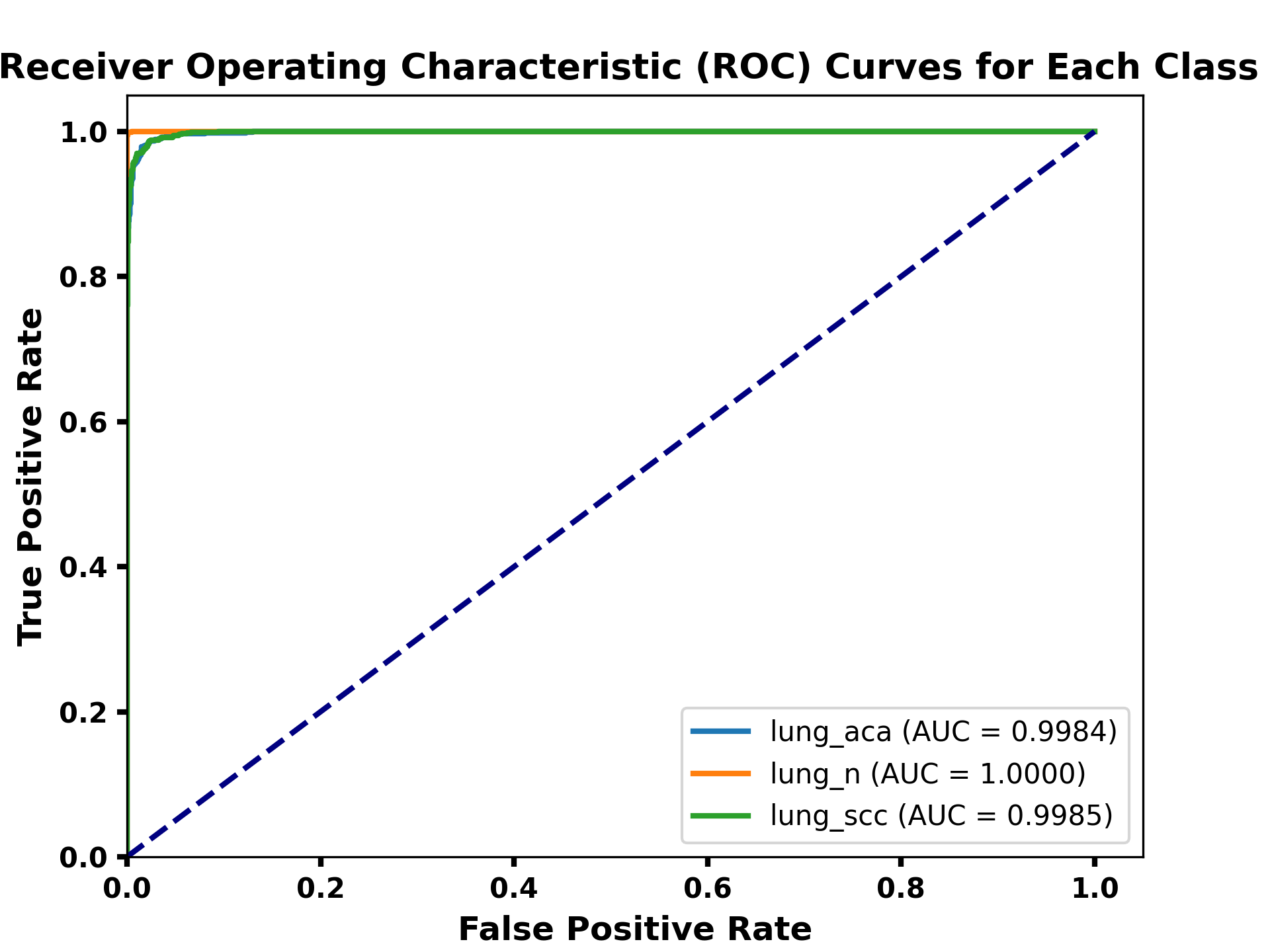}
  \vspace{-0.5cm} % Adds some space between the image and text
    \begin{center}
    \textbf{(b)}    
    \end{center}
\end{minipage}
  \caption{Using the LC25000 Dataset, the proposed model classwise performance is compared using the (a) P-R curve and the (b) ROC curve.}
  \label{fig:15.png}
\end{figure*}
Fig 9, PR and ROC curves offer a detailed evaluation of proposed FuzzyDistillViT-MobileNet model performance on the Histopathological CT-SCAN IQOTH/NCCD dataset. In the PR curve, all three classes (Benign, Malignant, and Normal) show very high precision and recall, with the Normal class achieving an efficient AU the PR Curve (AP = 1.0000), indicating that the model performs exceptionally well in identifying normal instances. The Malignant class follows closely, with a high AP of 0.9998, and the Benign class, while still showing robust performance with an AP of 0.9968, appears slightly less sharp in terms of recall and precision at higher recall rates. The ROC curve analysis further confirms these findings. The model achieves nearly perfect performance in distinguishing between the classes, with the Normal class reaching an AUC of 1.0000, which denotes perfect classification. The Malignant class follows with a very high AUC of 0.9998, and the Benign class has a slightly lower AUC of 0.9966. These AUC values indicate excellent overall model performance across all classes, with the model being particularly adept at differentiating between the Normal and Malignant classes. Comparing the PR and ROC curves highlights the consistency of the model performance across these two-evaluation metrics, underscoring the robustness and accuracy of your FuzzyDistillViT-MobileNet model for this dataset.
\begin{figure*}[h]
\centering
\begin{minipage}[]{6cm}
  \centering
  \includegraphics[width = 5cm]{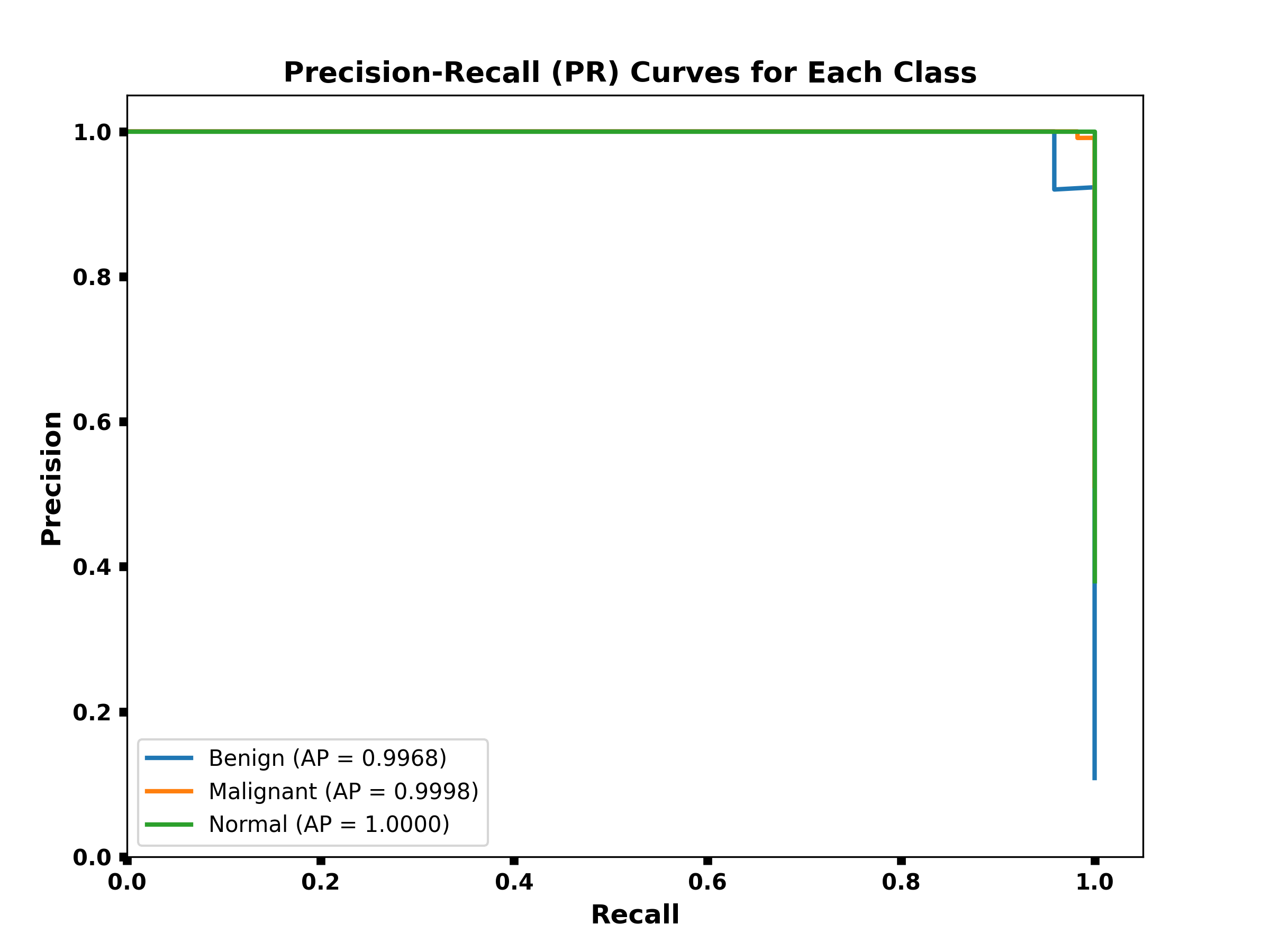}
  \vspace{-0.5cm} % Adds some space between the image and text
    \begin{center}
    \textbf{(a)}    
    \end{center}
\end{minipage}
\begin{minipage}[]{6cm}
  \centering
  \includegraphics[width = 5cm]{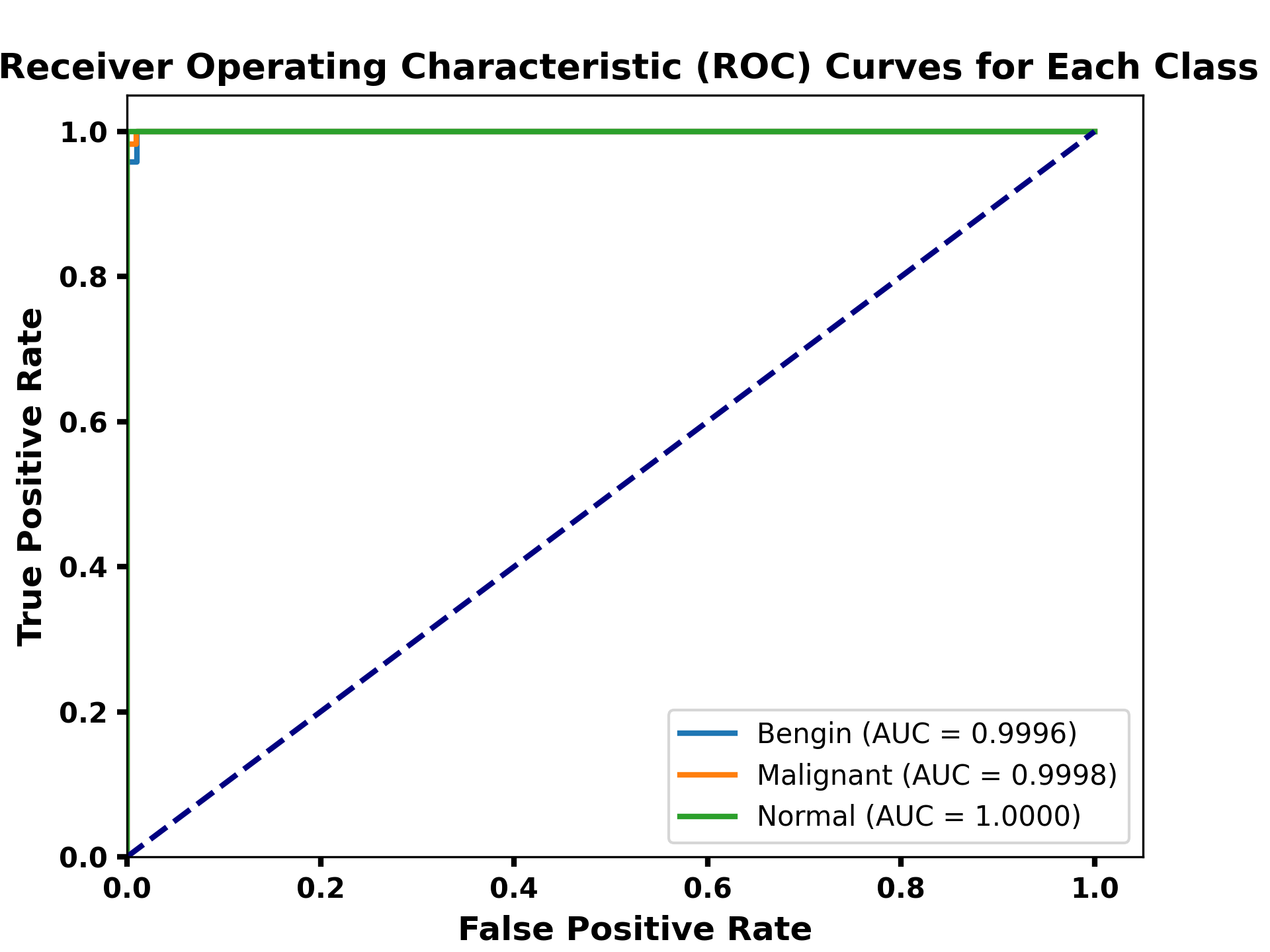}
  \vspace{-0.5cm} % Adds some space between the image and text
    \begin{center}
    \textbf{(b)}    
    \end{center}
\end{minipage}
  \caption{Using the IQOTH/NCCD Dataset, the proposed model classwise performance is compared using the (a) P-R curve and the (b) ROC curve.}
  \label{fig:15.png}
\end{figure*}
\subsection{Performance evaluation of the Scaled Weights (Histopathological LC25000): SOTA CNN}
Table 4 showcases the performance metrics for various DL models on a given dataset, highlighting the FuzzyDistillViT-MobileNet model as the best-performing model. It achieves an Accuracy of 99.16\%, Precision of 99.26\%, Recall of 98.95\%, and an F1-Score of 98.86\%, making it the top candidate among all models tested. These results emphasize the model excellent ability to correctly classify instances while maintaining a high degree of balance between false positives and false negatives, as reflected by its high precision and recall scores. DenseNet169 comes closest to the proposed model with an Accuracy of 98.66\%, Precision of 98.67\%, and F1-Score of 98.66\%, demonstrating its strong performance in comparison to other traditional DL architectures. MobileNetV2 also performs well with an Accuracy of 98.16\% and F1-Score of 98.16\%, positioning it as another competitive model. Other models like ResNet50V2 and VGG16 achieve solid performance with Accuracy around 97.73\%, but they still fall short of the proposed model’s superior results. On the lower end, models such as MobileNetV3Small and MobileNetV3Large exhibit significantly lower performance, with accuracies of 75.26\% and 80.36\%, respectively. The Proposed FuzzyDistillViT-MobileNet outperforms other models across all four metrics, demonstrating its effectiveness and ability to provide highly accurate predictions for the given task. This performance underscores the importance of using advanced architectures and hybrid models like FuzzyDistillViT-MobileNet for improving classification tasks, particularly in complex domains where accuracy, precision, recall, and F1-Score are critical. The results also show that models with more complex structures, such as DenseNet and MobileNet, tend to perform better than simpler models, while lighter architectures like MobileNetV3Small have notable limitations in handling the complexity of the dataset.
\begin{table}[h!]
\centering
\caption{Performance Comparison of FuzzyDistillViT-MobileNet and DL Models on the Histopathological LC25000 Dataset}
\begin{tabular}{lcccc}
\hline
\textbf{Methods} & \textbf{Accuracy (\%)} & \textbf{Precision (\%)} & \textbf{Recall (\%)} & \textbf{F1-Score (\%)} \\
\hline
ConvNeXtBase \cite{liu2022convnet} & 95.53 & 95.54 & 95.53 & 95.53 \\
ConvNeXtSmall \cite{liu2022convnet} & 85.90 & 85.95 & 85.90 & 85.84 \\
DenseNet121 \cite{huang2017densely} & 97.46 & 97.47 & 97.46 & 97.46 \\
DenseNet169 \cite{huang2017densely} & 98.66 & 98.67 & 98.66 & 98.66 \\
DenseNet201 \cite{huang2017densely} & 97.73 & 97.83 & 97.73 & 97.73 \\
InceptionV3 \cite{szegedy2016rethinking} & 95.50 & 95.65 & 95.50 & 95.51 \\
MobileNetV1 \cite{Howard2017Mobilenets} & 97.80 & 97.82 & 97.79 & 97.80 \\
MobileNetV2 \cite{Howard2017Mobilenets} & 98.16 & 98.17 & 98.16 & 98.16 \\
MobileNetV3Small \cite{howard2019searching} & 75.26 & 77.20 & 75.26 & 75.58 \\
MobileNetV3Large \cite{howard2019searching} & 80.36 & 80.19 & 80.36 & 80.03 \\
ResNet50V2 \cite{he2016deep} & 97.73 & 97.73 & 97.73 & 97.73 \\
ResNet101V2 \cite{he2016deep} & 97.13 & 97.25 & 97.13 & 97.12 \\
ResNet152V2 \cite{he2016deep} & 96.39 & 96.53 & 96.40 & 96.39 \\
VGG16 \cite{simonyan2014very} & 97.43 & 97.43 & 97.43 & 97.43 \\
VGG19 \cite{simonyan2014very} & 96.16 & 96.16 & 96.16 & 96.16 \\
VITB16 \cite{dosovitskiy2020image} & 97.00 & 97.00 & 97.00 & 97.01 \\
VITB32 \cite{dosovitskiy2020image} & 97.80 & 97.83 & 97.69 & 97.88 \\
VITL16 \cite{dosovitskiy2020image} & 96.10 & 96.28 & 96.10 & 96.09 \\
VITL32 \cite{dosovitskiy2020image} & 96.33 & 96.45 & 96.33 & 96.32 \\
Proposed (our) & 99.16 & 99.26 & 98.95 & 98.86 \\
\hline
\end{tabular}
\end{table}
\subsection{Performance evaluation of the Scaled Weights (CT-SCAN IQOTH/NCCD): SOTA CNN}
Table 5 compares the performance of various DL models across four metrics: Accuracy, Precision, Recall, and F1-Score. The FuzzyDistillViT-MobileNet model shows exceptional results, achieving the highest Accuracy (99.54\%), Precision (99.59\%), Recall (99.28\%), and F1-Score (99.36\%), indicating that it outperforms all other models across all metrics. This demonstrates the robustness and reliability of the proposed model in accurately classifying instances, with minimal errors. DenseNet169 follows closely with high scores, recording 98.66\% accuracy, 98.67\% precision, and 98.66\% recall, demonstrating its strong performance among the existing models, especially in comparison to DenseNet121 and DenseNet201, which show slightly lower scores. On the other hand, MobileNetV3Small and MobileNetV3Large exhibit relatively weaker performance, with accuracy values of 75.26\% and 80.36\%, respectively. Similarly, VITB16 and VITB32 also perform poorly compared to others, with accuracy percentages of 86.42\% and 89.59\%, respectively, and notably lower precision and recall rates, particularly in the VITB16 model. Models like ResNet50V2, MobileNetV2, and VGG16 perform quite well, with accuracy percentages ranging between 93.39\% to 97.80\%, indicating their strong capability in classification tasks but still falling short compared to the FuzzyDistillViT-MobileNet model. The overall trend shows that the FuzzyDistillViT-MobileNet model excels in all evaluation metrics, followed by DenseNet169 as a strong contender, while models such as MobileNetV3Small, VITB16, and VITB32 show suboptimal performance. These results underscore the superiority of the proposed model in this classification task.
\begin{table}[h!]
\centering
\caption{Performance Comparison of FuzzyDistillViT-MobileNet and DL models on the CT-SCAN IQOTH/NCCD Dataset}
\begin{tabular}{lcccc}
\hline
\textbf{Methods} & \textbf{Accuracy (\%)} & \textbf{Precision (\%)} & \textbf{Recall (\%)} & \textbf{F1-Score (\%)} \\
\hline
ConvNeXtBase \cite{liu2022convnet} & 95.53 & 95.54 & 95.53 & 95.53 \\
ConvNeXtSmall  \cite{liu2022convnet} & 85.90 & 85.95 & 85.90 & 85.84 \\
DenseNet121 \cite{huang2017densely} & 97.46 & 97.47 & 97.46 & 97.46 \\
DenseNet169 \cite{huang2017densely} & 98.66 & 98.67 & 98.66 & 98.66 \\
DenseNet201 \cite{huang2017densely} & 97.73 & 97.83 & 97.73 & 97.73 \\
InceptionV3 \cite{szegedy2016rethinking} & 95.50 & 95.65 & 95.50 & 95.51 \\
MobileNetV1 \cite{Howard2017Mobilenets} & 97.80 & 97.82 & 97.79 & 97.80 \\
MobileNetV2 \cite{Howard2017Mobilenets} & 97.43 & 98.16 & 97.88 & 97.65 \\
MobileNetV3Small \cite{howard2019searching} & 75.26 & 77.20 & 75.26 & 75.58 \\
MobileNetV3Large \cite{howard2019searching} & 80.36 & 80.19 & 80.36 & 80.03 \\
ResNet50V2 \cite{he2016deep} & 97.73 & 97.73 & 97.73 & 97.74 \\
ResNet101V2 \cite{he2016deep} & 97.13 & 97.25 & 97.13 & 97.12 \\
ResNet152V2 \cite{he2016deep} & 93.39 & 96.53 & 96.40 & 96.39 \\
VGG16 \cite{simonyan2014very} & 97.43 & 97.43 & 97.43 & 97.43 \\
VGG19 \cite{simonyan2014very} & 96.16 & 96.16 & 96.16 & 96.16 \\
VITB16 \cite{dosovitskiy2020image} & 86.42 & 74.57 & 71.53 & 72.34 \\
VITB32 \cite{dosovitskiy2020image} & 89.59 & 84.54 & 78.38 & 80.31 \\
VITL16 \cite{dosovitskiy2020image} & 88.23 & 85.43 & 76.78 & 79.54 \\
VITL32 \cite{dosovitskiy2020image} & 88.23 & 91.57 & 77.70 & 80.99 \\
Proposed (our) & 99.54 & 99.59 & 99.28 & 99.36 \\
\hline
\end{tabular}
\end{table}
\subsection{Performance evaluation of the Scaled Weights (Histogram + CT-SCAN): SOTA studies}
The FuzzyDistillViT-MobileNet model outperforms existing SOTA models on both the LC25000 and IQ-OTH/NCCD datasets, as demonstrated by its high accuracy of 99.16\% and 99.54\%, respectively. In comparison to other models like ResNet + Attention (89.80\%), SMA optimized (95.00\%), and VER-Net (91.00\%) on the LC25000 dataset, our model achieves significantly higher accuracy, highlighting its superior ability to handle complex medical images. Similarly, on the IQ-OTH/NCCD dataset, where models like Deep CNN (98.83\%) and Lung-EffNet (99.09\%) show strong performance, the FuzzyDistillViT-MobileNet model surpasses these approaches with an accuracy of 99.54\%. This demonstrates its robust performance across diverse imaging domains. The success of our proposed model can be attributed to several key innovations. The dynamic fuzzy logic-driven KD enables the model to adapt to varying levels of uncertainty in different regions of the medical images, allowing it to focus on high-confidence areas and improve classification accuracy. Additionally, the use of ViT-B32 as the instructor model and MobileNet as the student model, coupled with the GA for selecting the optimal teacher model, contributes to a more efficient and accurate training process. These features, combined with pixel-level image improvement techniques like Gamma correction and Histogram Equalization, enhance the model's ability to learn complex features and outperform other models, particularly in the challenging task of medical image classification.  Table 6 presents the detail performance evaluation of the proposed model with SOTA studies.
\begin{table}[h!]
\centering
\caption{Performance Comparison of FuzzyDistillViT-MobileNet and SOTA studies on the Histogram/CT-SCAN IQOTH/NCCD Dataset}
\begin{tabular}{cccc}
\hline
\textbf{Reference} & \textbf{Dataset} & \textbf{Methods} & \textbf{Accuracy} \\
\hline
Zhang et al. \cite{Zhang2023} & LC25000 & ResNet + Attention & 89.80\% \\
Nagaraj et al. \cite{Nagaraj2025} & LC25000 & SMA optimized & 95.00\% \\
Saha et al. \cite{Saha2024} & LC25000 & VER-Net & 91.00\% \\
Our (Proposed) & LC25000 & FuzzyDistillViT-MobileNet & 99.16\% \\
Humayun et al. \cite{Humayun2022} & IQ-OTH/NCCD & Deep CNN model & 98.83\% \\
Yan et al. \cite{Yan2023} & IQ-OTH/NCCD & SOA-CNN & 96.58\% \\
Raza et al. \cite{Raza2023} & IQ-OTH/NCCD & Lung-EffNet & 99.09\% \\
Mehrzadi et al. \cite{Mehrzadi2025} & IQ-OTH/NCCD & CNN-model & 98.32\% \\
Proposed (our) & IQ-OTH/NCCD & FuzzyDistillViT-MobileNet & 99.54\% \\
\hline
\end{tabular}
\label{table:comparison}
\end{table}
\subsection{Interpretability and visualization analysis: FuzzyDistillViT-MobileNet}
Section 4.7 focuses on the interpretability and visualization analysis of the FuzzyDistillViT-MobileNet model, aiming to enhance the understanding of its decision-making process. This section includes two key techniques: GRADCAM and GRADCAM++, which provide insights into the model attention and feature importance by visualizing the regions that contribute most to the predictions. Additionally, LIME (Local Interpretable Model-agnostic Explanations) analysis is employed to offer local explanations for individual predictions, further elucidating the model behavior and boosting its transparency. These methods collectively support model interpretability, ensuring its reliability and helping to uncover underlying decision patterns.
\subsubsection{GRADCAM and GRADCAM++}
In the FuzzyDistillViT-MobileNet framework, both GRAD-CAM and GRAD-CAM++ are applied to the student model (MobileNet), visualizing which regions in the input images are most influential in making predictions. The generated heatmaps highlight areas of the image that are critical for classification, aiding in the interpretability of the model. In the heatmaps produced by GRAD-CAM, the regions that strongly influence the decision are marked with warm colors (e.g., red), whereas the less significant areas are shown in cooler colors (e.g., blue). These heatmaps are overlaid on the original images, allowing medical experts to visually confirm that the model focuses on the correct parts of the image. 

   The GRAD-CAM++ heatmaps offer a more refined visualization. Unlike GRAD-CAM, which aggregates gradients linearly, GRAD-CAM++ uses higher-order gradients for more detailed and precise localization of critical features. This refinement results in sharper, more focused heatmaps, which can capture subtle patterns in the data, especially useful for tasks requiring high precision, such as medical image classification. The GRAD-CAM++ method produces heatmaps that allow for a finer distinction of relevant features, making it easier to detect smaller or more complex structures that contribute to the classification decision. The heatmaps generated by both GRAD-CAM and GRAD-CAM++ in the FuzzyDistillViT-MobileNet model not only provide transparency into the decision-making process but also validate the effectiveness of KD between the ViT instructor model and the MobileNet student model. These visualizations ensure that the student model learns to focus on the same key areas that the instructor model highlights, thus ensuring the success of the distillation process. The overlaid heatmaps confirm that the model’s attention is appropriately directed towards clinically significant regions in the images, such as tumor areas in histopathological slides or critical structures in CT scans, further enhancing the trust and interpretability of the model predictions. As shown in Fig 10, both GRAD-CAM and GRAD-CAM++ visualizations reveal the critical regions in the histopathological and CT-scan images that allowing for a finer-grained understanding of the model attention on relevant image areas.
\begin{figure*}[h]
\centering
\begin{minipage}[]{7cm}
  \centering
  \includegraphics[width = 7cm]{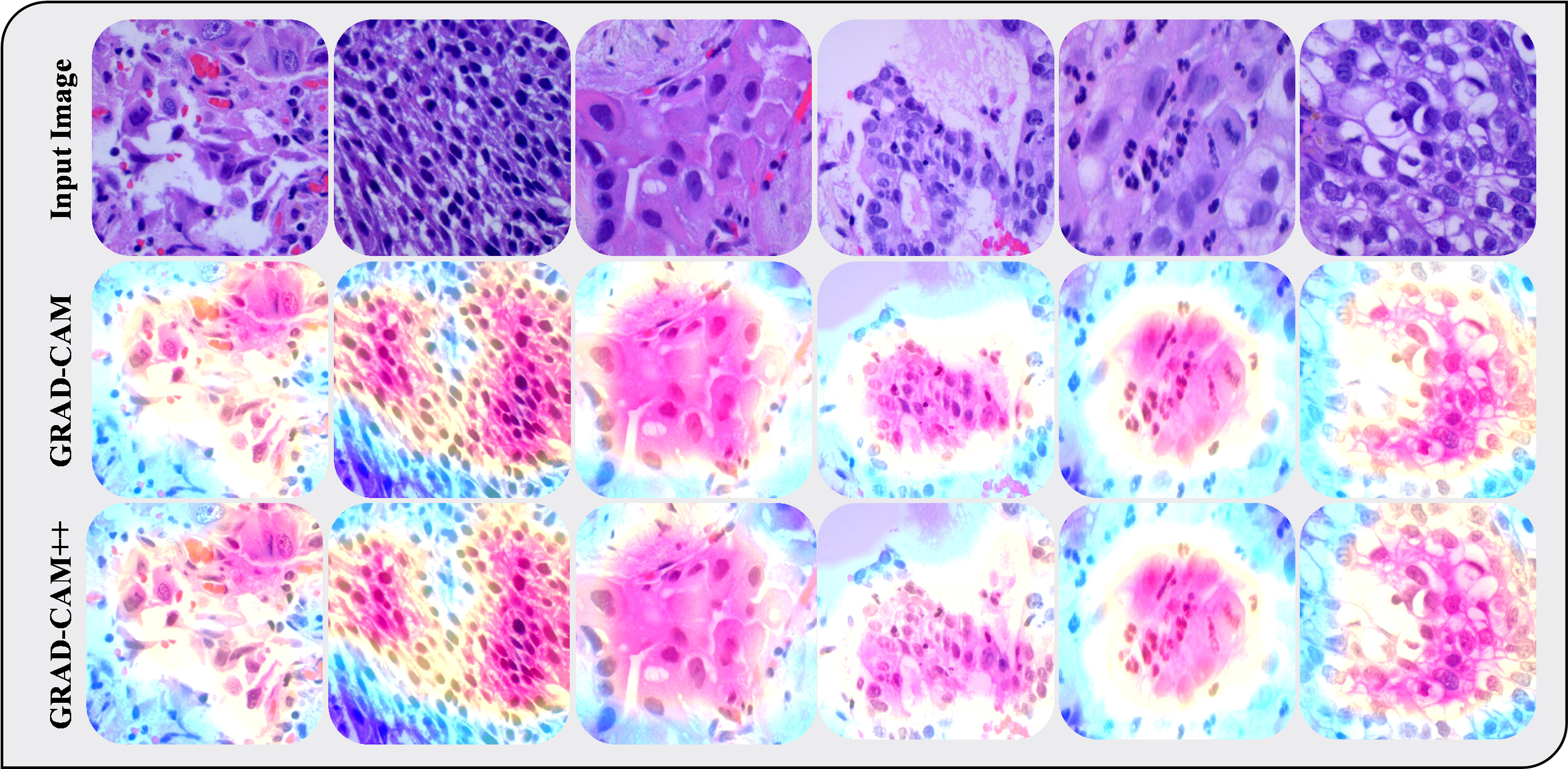}
  \vspace{-0.5cm} % Adds some space between the image and text
    \begin{center}
    \textbf{(a) LC25000 (Histopathological LC)}    
    \end{center}
\end{minipage}
\begin{minipage}[]{7cm}
  \centering
  \includegraphics[width = 7cm]{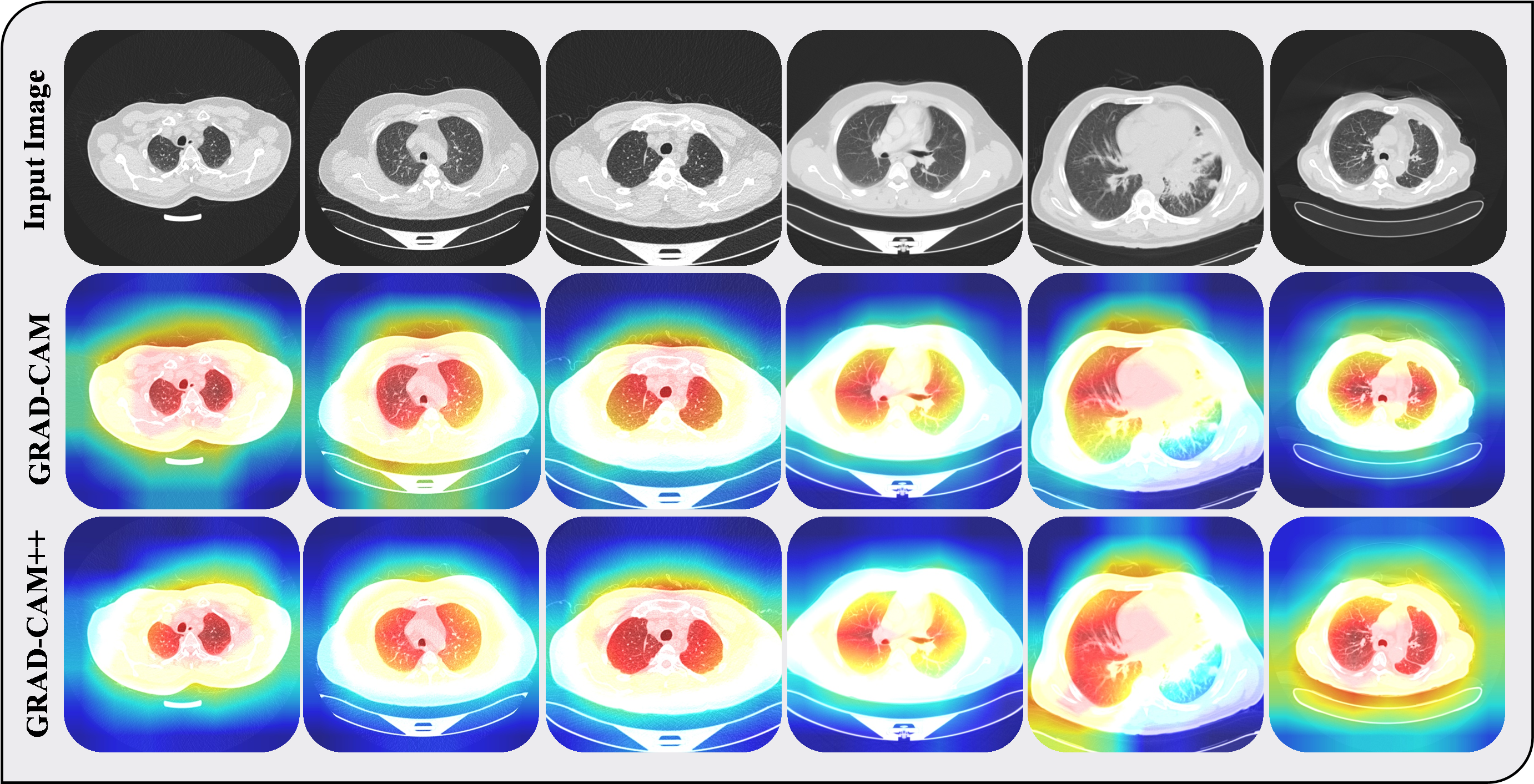}
  \vspace{-0.5cm} % Adds some space between the image and text
    \begin{center}
    \textbf{(b) IQOTH/NCCD (CT-SCAN LC)}    
    \end{center}
\end{minipage}
  \caption{Heatmap Visualizations of the MobileNet Student model using GRAD-CAM and GRAD-CAM++ on Histopathological and CT-Scan Images. (a) LC25000 dataset and (b) GRAD-CAM heatmap for CT-scan images from the IQOTH/NCCD dataset, highlighting the key areas contributing to the classification decision}
  \label{fig:15.png}
\end{figure*}
\subsubsection{LIME analysis}
Fig 11, image showcases the application of LIME (Local Interpretable Model-agnostic Explanations) to the MobileNet student model in the FuzzyDistillViT-MobileNet framework. LIME generates local explanations for the model predictions by approximating the decision-making process with an interpretable surrogate model. In the visualizations, the input images (histopathological and CT-scan images) are accompanied by LIME heatmaps that highlight the regions most influential in the model decision. The areas with higher importance are marked in warmer colors (such as red), while the regions with lower significance are shown in cooler colors (green or blue). The LIME technique provides transparency by showing exactly which regions of the image the model is attending to when making predictions. For example, in histopathological images, LIME emphasizes the critical areas of tissue that contribute to the classification, such as potential cancerous cells. Similarly, for CT-scan images, LIME identifies the key regions, like tumors or other significant structures. This local interpretability helps to ensure that the MobileNet student model is making decisions based on relevant features, increasing the trustworthiness and clinical usability of the model.
\begin{figure}[h!]
    \centering
    \includegraphics[width = 9cm]{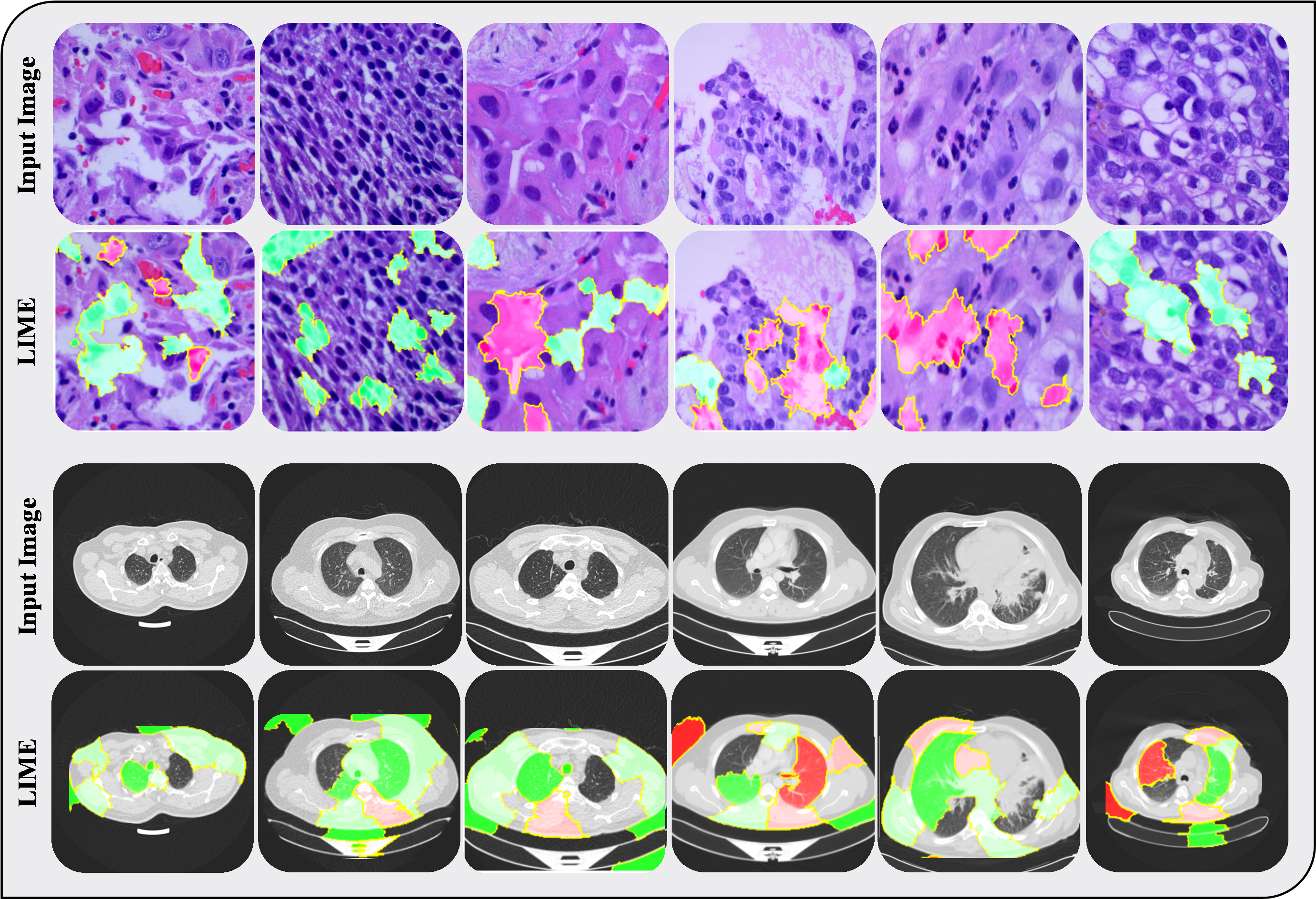}
    \caption{LIME Explanations for the MobileNet Student Model Predictions on Histopathological and CT-Scan Images. The LIME heatmaps illustrate the key regions in both histopathological LC images and CT-scan images that contribute to the MobileNet student model's predictions.}
    \label{fig:se.png}
\end{figure}
\subsection{Real-time FuzzyDistillViT-MobileNet testing: Android application}
Validating model performance and making sure they can manage the intricacies and variances present in actual clinical circumstances need testing them on real-time datasets. Developing predictive models for lung cancer diagnosis is crucial in the medical industry in order to diagnose and prevent cancer early. We can evaluate the model's resilience and flexibility by running real-time tests, which will produce more accurate predictions in real-world scenarios.

We have created an Android app that uses our FuzzyDistillViT-MobileNet LC prediction model in real time.  The application uses two data modalities and image processing algorithms-2 to reliably detect and differentiate between several LC diseases. Our real-time approach allows the app to provide accurate results, improving the medical professional inspection process and offering a dependable clinical application tool.  The efficient accuracy of the model guarantees that it can identify LC early on, enhancing the effectiveness of disease treatment.  For on-site use, this solution provides a smooth, user-friendly platform.  The proposed FuzzyDistillViT-MobileNet model is evaluated in real time via an Android application, as shown in Fig 12.
\begin{figure}[h!]
    \centering
    \includegraphics[width = 9cm]{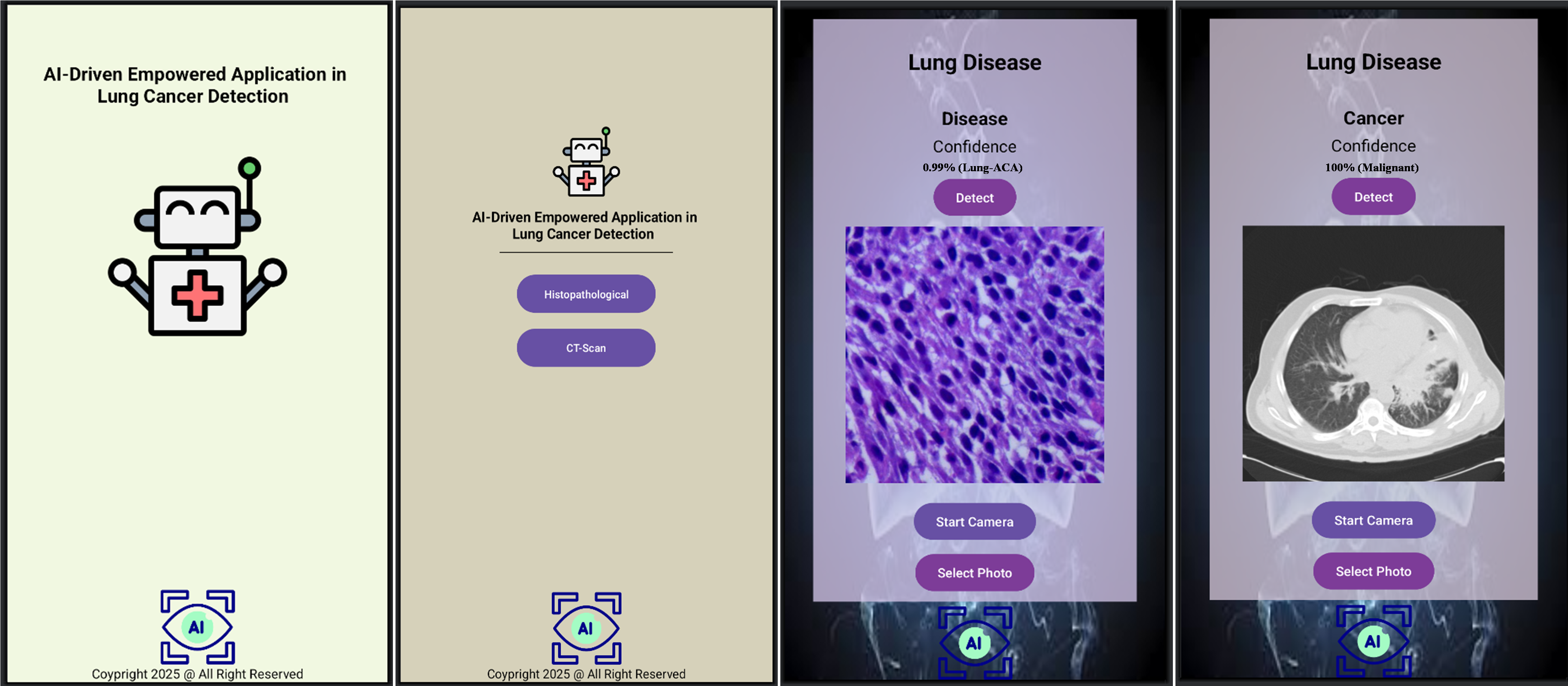}
    \caption{Overview of Real-time FuzzyDistillViT-MobileNet testing: Android application}
    \label{fig:se.png}
\end{figure}
\subsubsection*{Target Device}
A device running Android 5.5 or later and having at least 4GB of RAM is the minimum hardware requirement for the Android application created to apply our real-time LC detection technique.  We noticed certain limits during real-world testing, such as sporadic lags when processing huge image files and variations in performance on systems with less RAM.  Notwithstanding these difficulties, the software was able to accurately detect LC; nevertheless, device specs may affect performance.  To preserve real-time functioning on devices with less computing power, the image processing algorithms-2 also needed to be optimized.
\subsubsection*{Design Step}
The steps necessary to install DL models on Android smartphones. Keras can be considered the primary prototyping library due to its ability to extract TFLite for Android and its ease of conversion to the TensorFlow backend (FuzzyDistillViT-MobileNet) model. The FuzzyDistillViT-MobileNet model is kept as inference only, and all training-related layers are removed, allowing only the feed-forward path of a network model to operate. Because layers are maintained as computational graphs, they can be tailored to the platform on which they will function.

The process of deploying the FuzzyDistillViT-MobileNet model to Android devices involves several key steps to ensure the model is optimized and functional on resource-constrained environments such as smartphones. Initially, the model is loaded and trained using standard TensorFlow methods. During the training phase, the model learns to improve its accuracy and performance on the LC prediction. Once training is complete, all variables within the model are converted using TensorFlow’s Graph-util submodule. This step ensures that the model parameters are compatible with the next stages of conversion and deployment.

Next, the trained model is saved in the .h5 format, which is a widely accepted file format for storing TensorFlow models. This format ensures that the model architecture, weights, and training configurations are preserved for future use. Following this, the model undergoes a conversion process to TensorFlow Lite (TFLite), a framework specifically designed for optimizing DL models for mobile and embedded devices. The conversion to TFLite reduces the model size and optimizes its operations for efficient execution on mobile hardware. The model is also quantized during this process, reducing its memory footprint and computational complexity without sacrificing significant performance.

Finally, the converted TensorFlow Lite model is deployed to Android cell phones. By running the model on Android devices, it enables real-time inference directly on the smartphone, eliminating the need for constant cloud communication and reducing latency. This deployment ensures that the model can perform LC prediction efficiently even in the absence of powerful computing resources. The entire process enhances the accessibility and usability of DL models in mobile applications, allowing for on-device processing, faster response times, and improved user experience without requiring extensive cloud infrastructure.
\subsection{Ablation study}
Understanding the contribution of each component within a model requires ablation studies. As shown in Table 7, we execute ablation research in this part to evaluate the effect of Fusion Pixel 1\textsuperscript{st}   with Pixel 2\textsuperscript{nd}  Level by Level on the student model performance. Table 7 presents a performance comparison of the FuzzyDistillViT-MobileNet model across different pixel-level configurations, evaluating its accuracy, precision, recall, and F1-score. The results show that for both LC25000 and IQOTH/NCCD methods, combining the 1\textsuperscript{st} and 2\textsuperscript{nd} pixel levels generally improves performance compared to using a single pixel level. Specifically, the LC25000 method achieves its highest performance with the combination of Pixel 1\textsuperscript{st}  + Pixel 2\textsuperscript{nd}, reaching 99.16\% accuracy and 98.86\% F1-score. Similarly, the IQOTH/NCCD method also performs best with the combined configuration, achieving 99.54\% accuracy and 99.36\% F1-score. The improvements in precision, recall, and F1-score are notable, particularly with the Pixel 2\textsuperscript{nd} and combined configurations for both methods, suggesting that incorporating additional pixel-level data enhances the model’s performance across various metrics.
\begin{table}[h!]
\centering
\caption{Performance Comparison of FuzzyDistillViT-MobileNet with Different Pixel Level Configuration}
\begin{tabular}{llcccc}
\hline
\textbf{Methods} & \textbf{Levels} & \textbf{Accuracy (\%)} & \textbf{Precision (\%)} & \textbf{Recall (\%)} & \textbf{F1-Score (\%)} \\
\hline
LC25000 & Pixel 1\textsuperscript{st} & 98.16 & 98.17 & 98.16 & 98.16 \\
 & Pixel 2\textsuperscript{nd} & 98.66 & 98.67 & 98.66 & 98.66 \\
 & Pixel 1\textsuperscript{st} + Pixel 2\textsuperscript{nd} & 99.16 & 99.26 & 98.95 & 98.86 \\
\hline
IQOTH/NCCD & Pixel 1\textsuperscript{st} & 98.64 & 98.40 & 98.88 & 98.43 \\
 & Pixel 2\textsuperscript{nd} & 99.09 & 99.67 & 99.60 & 98.73 \\
 & Pixel 1\textsuperscript{st} + Pixel 2\textsuperscript{nd} & 99.54 & 99.59 & 99.28 & 99.36 \\
\hline
\end{tabular}
\end{table}
Table 8 presents the performance comparison of the FuzzyDistillViT-MobileNet model with various instructor configurations across two datasets: LC25000 and IQOTH/NCCD. The evaluation metrics include Precision, Recall, F1-Score, and Accuracy. For the LC25000 dataset, the ViTB16 + MobileNet configuration achieves the highest performance with a Precision of 99.09\%, Recall of 99.63\%, F1-Score of 99.70\%, and Accuracy of 98.88\%, while other configurations with larger ViT models (ViTB32, ViTL16, ViTL32) show slightly lower performance. Similarly, for the IQOTH/NCCD dataset, the ViTB32 + MobileNet configuration performs best with a Precision of 99.54\%, Recall of 99.59\%, and an F1-Score of 99.28\%, resulting in an Accuracy of 99.36\%. Overall, the ViTB16 + MobileNet and ViTB32 + MobileNet configurations consistently outperform others, indicating their superior ability to balance the metrics across both datasets.
\begin{table}[h!]
\centering
\caption{Performance Comparison of FuzzyDistillViT-MobileNet with different Instructor configuration}
\begin{tabular}{cccccc}
\hline
\textbf{Dataset} & \textbf{Instructor} & \textbf{Precision} & \textbf{Recall} & \textbf{F1-Score} & \textbf{Accuracy} \\
\hline
\multirow{4}{*}{LC25000} & ViTB16 + MobileNet & 99.09 & 99.63 & 99.70 & 98.88 \\
                          & ViTB32 + MobileNet & 99.16 & 99.26 & 98.95 & 98.86 \\
                          & ViTL16 + MobileNet & 97.43 & 97.43 & 97.43 & 97.43 \\
                          & ViTL32 + MobileNet & 97.00 & 97.00 & 97.00 & 97.01 \\
\multirow{4}{*}{IQOTH/NCCD} & ViTB16 + MobileNet & 98.66 & 98.67 & 98.66 & 98.66 \\
                           & ViTB32 + MobileNet & 99.54 & 99.59 & 99.28 & 99.36 \\
                           & ViTL16 + MobileNet & 97.73 & 97.73 & 97.73 & 97.74 \\
                           & ViTL32 + MobileNet & 97.80 & 97.82 & 97.79 & 97.80 \\
\hline
\end{tabular}
\end{table}
\section{Conclusion}
In this paper, we introduced the FuzzyDistillViT-MobileNet model, a cutting-edge approach for LC image classification that leverages dynamic fuzzy logic-driven KD to address the complexities and uncertainties inherent in disease diagnosis. By dynamically adjusting distillation weights using fuzzy logic, our model enables the student model (MobileNet) to focus more on high-confidence regions and reduce attention to ambiguous areas, significantly enhancing performance in challenging LC image analysis. The use of ViT-B32 as the instructor model, combined with a dynamic wait adjustment mechanism, ensures the model ability to capture long-range dependencies and improve convergence. We further optimize the selection of the most suitable pre-trained student model through the application of GA, which outperforms traditional optimizers by exploring a wider solution space and avoiding local optima. This ensures that the best model is selected, balancing computational cost and performance. To enhance the quality of input images, we employed Pixel-level image improvement techniques such as Gamma correction and Histogram Equalization. The images were then fused using a wavelet-based fusion method, improving image resolution and preserving key features across multiple scales. The proposed model achieved outstanding results on diverse medical imaging datasets, including LC25000 histopathological images (99.16\% accuracy) and IQOTH/NCCD CT-scan images (99.54\% accuracy), demonstrating its robustness and ability to generalize across multiple medical domains. Moreover, GRAD-CAM, GRAD-CAM++, and LIME were utilized to provide interpretability, ensuring transparency and trust in the model’s decision-making process. Through rigorous comparative analysis, the model performance was validated against SOTA methods, further confirming its competitiveness and ability to handle uncertainty in real-world scenarios. Additionally, the deployment of the model via an Android application for real-time use in medical environments demonstrates its practical applicability. The FuzzyDistillViT-MobileNet model, with its combination of high accuracy, image improvement techniques, and interpretability, represents a valuable advancement in the field of medical image analysis and has the potential to significantly impact clinical practices by aiding in the diagnosis of complex diseases.
  
  Future work will focus on expanding the model capabilities to handle additional imaging modalities and integrating multi-modal data to enhance diagnostic accuracy. Additionally, we plan to further optimize the model for deployment in resource-constrained environments, ensuring its practical usability in a broader range of healthcare settings.

\backmatter
\section*{Declarations}
\bmhead{Ethics approval and consent to participate}
Approved and Not applicable
\bmhead{Consent for publication}
Not applicable
\bmhead{Data availability}
Data will be made available on request.
\bmhead{Funding}
No funding
\bmhead{Declaration of competing interest}
The authors declare that they have no known competing financial interests or personal relationships that could have appeared to influence the work reported in this paper.
\bmhead{CRediT authorship contribution statement}
Saif Ur Rehman Khan \& Muhammed Nabeel Asim: Conceptualization, Data curation, Methodology, Software, Validation, Writing original draft \& Formal analysis. Sebastian Vollmer: Conceptualization, Funding acquisition, Supervision. Andreas Dengel: review \& editing. 
\bibliography{sn-bibliography}% common bib file
\end{document}